\documentclass{article}

\usepackage{url}            
\usepackage{times}
\usepackage{epsfig}
\usepackage{graphicx}
\usepackage{amsmath}
\usepackage{amssymb}
\usepackage{arydshln}
\usepackage{multirow,bigdelim}
\usepackage{booktabs}
\usepackage[dvipsnames,table]{xcolor}
\usepackage{multicol}
\usepackage{tabularx}
\usepackage{diagbox}

\usepackage{booktabs}

\usepackage{colortbl}
\usepackage[breaklinks=true,bookmarks=false]{hyperref}

\usepackage[accepted]{icml2021}

\newcommand\blfootnote[1]{%
  \begingroup
  \renewcommand\thefootnote{}\footnote{#1}%
  \addtocounter{footnote}{-1}%
  \endgroup
}

\begin{document}
\setlength{\footskip}{40pt}


\twocolumn[
\icmltitle{Revisiting ResNets: Improved Training and Scaling Strategies}

\begin{icmlauthorlist}
\icmlauthor{Irwan Bello}{goog}
\icmlauthor{William Fedus}{goog}
\icmlauthor{Xianzhi Du}{goog}
\icmlauthor{Ekin D. Cubuk}{goog}
\icmlauthor{Aravind Srinivas}{berk}
\icmlauthor{Tsung-Yi Lin}{goog}
\icmlauthor{Jonathon Shlens}{goog}
\icmlauthor{Barret Zoph}{goog}
\end{icmlauthorlist}

\icmlaffiliation{goog}{Google Brain}
\icmlaffiliation{berk}{UC Berkeley}
\icmlcorrespondingauthor{Irwan Bello and Barret Zoph}{\{ibello,barretzoph\}@google.com}

\vskip 0.3in
]

\printAffiliationsAndNotice{}

\begin{abstract}
Novel computer vision architectures monopolize the spotlight, but the impact of the model architecture is often conflated with simultaneous changes to training methodology and scaling strategies.
Our work revisits the canonical ResNet~\cite{resnet} and studies these three aspects in an effort to disentangle them.
Perhaps surprisingly, we find that training and scaling strategies may matter more than architectural changes, and further, that the resulting ResNets match recent state-of-the-art models.
We show that the best performing scaling strategy depends on the training regime and offer two new scaling strategies: (1) scale model depth in regimes where overfitting can occur (width scaling is preferable otherwise); (2) increase image resolution more slowly than previously recommended~\cite{tan2019efficientnet}.
Using improved training and scaling strategies, we design a family of ResNet architectures, ResNet-RS, which are 1.7x - 2.7x faster than EfficientNets on TPUs, while achieving similar accuracies on ImageNet.
In a large-scale semi-supervised learning setup, ResNet-RS achieves 86.2\% top-1 ImageNet accuracy, while being 4.7x faster than EfficientNet-NoisyStudent.
The training techniques improve transfer performance on a suite of downstream tasks (rivaling state-of-the-art self-supervised algorithms) and extend to video classification on Kinetics-400.
We recommend practitioners use these simple revised ResNets as baselines for future research.
\end{abstract}
\section{Introduction}
The performance of a vision model is a product of the architecture, training methods and scaling strategy.
However, research often emphasizes architectural changes.
Novel architectures underlie many advances, but are often simultaneously introduced with other critical -- and less publicized -- changes in the details of the training methodology and hyperparameters.
Additionally, new architectures enhanced by modern training methods are sometimes compared to older architectures with dated training methods (e.g. ResNet-50 with ImageNet Top-1 accuracy of 76.5\%~\cite{resnet}).
Our work addresses these issues and empirically studies the impact of \emph{training methods} and \emph{scaling strategies} on the popular ResNet architecture~\cite{resnet}.

\begin{figure}[t!]
    \begin{center}
    \includegraphics[width=\linewidth]{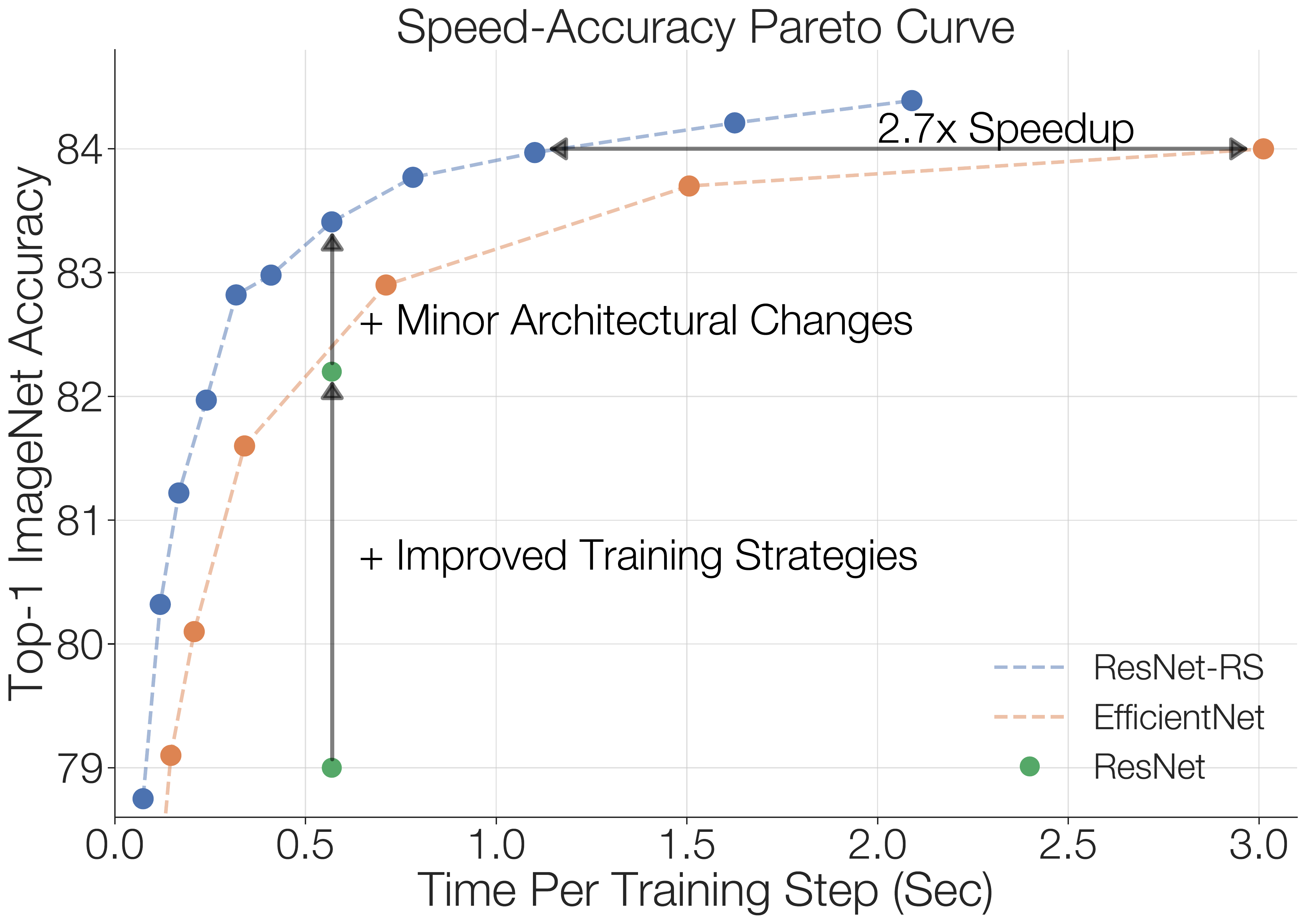}
    \end{center}
    \vspace{-0.3cm}
    \caption{\textbf{Improving ResNets to state-of-the-art performance.}
    We improve on the canonical ResNet~\cite{resnet} with modern training methods (as also used in EfficientNets~\cite{tan2019efficientnet}), minor architectural changes and improved scaling strategies.  
    The resulting models, \textbf{ResNet-RS}, outperform EfficientNets on the speed-accuracy Pareto curve with speed-ups ranging from \textbf{1.7x - 2.7x} on TPUs and \textbf{2.1x - 3.3x} on GPUs.
    ResNet (\textcolor{Green}{•}) is a ResNet-200 trained at 256$\times$256 resolution. 
    Training times reported on TPUs.
    }
    \label{fig:promo_figure}
    \vspace{-0.15cm}
\end{figure}

We survey the modern training and regularization techniques widely in use today and apply them to ResNets (Figure \ref{fig:promo_figure}).\blfootnote{* Code and checkpoints available in TensorFlow: \\\textcolor{magenta}{ \url{https://github.com/tensorflow/models/tree/master/official/vision/beta}} \\ \url{https://github.com/tensorflow/tpu/tree/master/models/official/resnet/resnet_rs}}{}
In the process, we encounter interactions between training methods and show a benefit of reducing weight decay values when used in tandem with other regularization techniques. 
An additive study of training methods in Table \ref{tab:resnet_method_ablation} reveals the significant impact of these decisions: a canonical ResNet with 79.0\% top-1 ImageNet accuracy is improved to 82.2\% (\textcolor{blue}{+3.2\%}) through \emph{improved training methods alone}.
This is increased further to 83.4\% by two small and commonly used architectural improvements: ResNet-D~\cite{he2019bag} and Squeeze-and-Excitation~\cite{hu2018squeeze}.
Figure \ref{fig:promo_figure} traces this refinement over the starting ResNet in a speed-accuracy Pareto curve.

We offer new perspectives and practical advice on scaling vision architectures.
While prior works extrapolate scaling rules from small models~\cite{tan2019efficientnet} or from training for a small number of epochs~\cite{radosavovic2020designing}, we design scaling strategies by exhaustively training models across a variety of scales for the full training duration (e.g. 350 epochs instead of 10 epochs).
In doing so, we uncover strong dependencies between the best performing scaling strategy and the training regime (e.g. number of epochs, model size, dataset size). 
These dependencies are missed in any of these smaller regimes, leading to sub-optimal scaling decisions.
Our analysis leads to new \emph{scaling strategies} summarized as \textbf{(1)} scale the model depth when overfitting can occur (scaling the width is preferable otherwise) and \textbf{(2)} scale the image resolution more slowly than prior works~\cite{tan2019efficientnet}.

Using the improved training and scaling strategies, we design re-scaled ResNets, \emph{ResNet-RS}, which are trained across a wide range of model sizes, as shown in Figure~\ref{fig:promo_figure}.
ResNet-RS models use less memory during training and are \textbf{1.7x - 2.7x} faster on TPUs (\textbf{2.1x - 3.3x} faster on GPUs) than the popular EfficientNets on the speed-accuracy Pareto curve. 
In a large-scale semi-supervised learning setup, ResNet-RS obtains a \textbf{4.7x} training speed-up on TPUs (\textbf{5.5x} on GPUs) over EfficientNet-B5 when co-trained on ImageNet and an additional 130M pseudo-labeled images.

Finally, we conclude with a suite of experiments testing the generality of the improved training and scaling strategies.
We first design a faster version of EfficientNet using our scaling strategy, \emph{EfficientNet-RS}, which improves over the original on the speed-accuracy Pareto curve.
Next, we show that the improved training strategies yield representations that rival or outperform those from self-supervised algorithms (SimCLR~\cite{chen2020simple} and SimCLRv2~\cite{chen2020big}) on a suite of downstream tasks.
The improved training strategies extend to video classification as well.
Applying the training strategies to 3D-ResNets on the Kinetics-400 dataset yields an improvement from 73.4\% to 77.4\% \textcolor{blue}{(+4.0\%)}.

Through combining minor architectural changes (used since 2018) and improved training and scaling strategies, we discover the ResNet architecture sets a state-of-the-art baseline for vision research. 
This finding highlights the importance of teasing apart each of these factors in order to understand what architectures perform better than others.

We summarize our contributions:
\begin{itemize}
    \itemsep0em
    \item An empirical study of regularization techniques and their interplay, which leads to a regularization strategy that achieves strong performance (+3\% top-1 accuracy) \emph{without having to change the model architecture}.
    \item A simple scaling strategy: (1) scale depth when overfitting can occur (scaling width can be preferable otherwise) and (2) scale the image resolution more slowly than prior works~\cite{tan2019efficientnet}. This scaling strategy improves the speed-accuracy Pareto curve of both ResNets and EfficientNets.
    \item \textbf{ResNet-RS}: a Pareto curve of ResNet architectures that are \textbf{1.7x - 2.7x} faster than EfficientNets on TPUs (\textbf{2.1x - 3.3x} on GPUs) by applying the training and scaling strategies.
    \item Semi-supervised training of ResNet-RS with an additional 130M pseudo-labeled images achieves 86.2\% top-1 ImageNet accuracy, while being \textbf{4.7x} faster on TPUs (\textbf{5.5x} on GPUs) than the corresponding EfficientNet-NoisyStudent~\cite{xie2020self}.
    \item ResNet checkpoints that, when fine-tuned on a diverse set of computer vision tasks, rival or outperform state-of-the-art self-supervised representations from SimCLR~\cite{chen2020simple} and SimCLRv2~\cite{chen2020big}.
    \item 3D ResNet-RS by extending our training methods and architectural changes to video classification. The resulted model improves the top-1 Kinetics-400 accuracy by \textbf{4.8\%} over the baseline. 
\end{itemize}
\section{Characterizing Improvements on ImageNet}
Since the breakthrough of AlexNet~\cite{krizhevsky2012imagenet} on ImageNet~\cite{ilsvrc}, a wide variety of improvements have been proposed to further advance image recognition performance.
These improvements broadly arise along four orthogonal axes: \emph{architecture, training/regularization methodology, scaling strategy and using additional training data}.

\paragraph{Architecture.}
The works that perhaps receive the most attention are novel architectures.
Notable proposals since AlexNet~\cite{krizhevsky2012imagenet} include VGG~\cite{simonyan2014very}, ResNet~\cite{resnet}, Inception~\cite{szegedy2015going,szegedy2016rethinking}, and ResNeXt~\cite{xie2017aggregated}.
Automated search strategies for designing architectures have further pushed the state-of-the-art, notably with NasNet-A~\cite{zoph2018learning}, AmoebaNet-A~\cite{real2019regularized} and EfficientNet~\cite{tan2019efficientnet}.
There have also been efforts in going beyond standard ConvNets for image classification, 
by adapting self-attention~\citep{vaswani2017attention} to the visual domain
~\cite{bello2019attention,ramachandran2019stand,hu2019local,shen2020global,dosovitskiy2020image}
or using alternatives such as lambda layers~\cite{bello2021lambdanetworks}.

\paragraph{Training and Regularization Methods.}
ImageNet progress has been boosted by innovations in training and regularization approaches.
When training models for more epochs, regularization methods such as dropout~\cite{srivastava2014dropout}, 
label smoothing~\cite{szegedy2016rethinking},
stochastic depth~\cite{huang2016deep},
dropblock~\cite{ghiasi2018dropblock}
and data augmentation~\cite{zhang2017mixup,yun2019cutmix,cubuk2018autoaugment,cubuk2019randaugment} have significantly improved generalization.
Improved learning rate schedules~\cite{loshchilov2016sgdr,goyal2017accurate} have further increased final accuracy.
While benchmarking architectures in a short non-regularized training setup facilitates fair comparisons with prior work, it is unclear whether architectural improvements are sustained at larger scales and improved training setups.
For example, the RegNet architecture~\cite{radosavovic2020designing} shows strong speedups over baselines in a short non-regularized training setup, but was not tested in a state-of-the-art ImageNet setup (best top-1 is 79.9\%).

\paragraph{Scaling Strategies.}
Increasing the model dimensions (e.g. width, depth and resolution) has been another successful axis to improve quality~\cite{rosenfeld2019constructive,hestness2017deep}.
Sheer scale was exhaustively demonstrated to improve performance of neural language models~\cite{kaplan2020scaling} which motivated the design of ever larger models including GPT-3~\cite{brown2020language} and Switch Transformer~\cite{fedus2021switch}.
Similarly, scale in computer vision has proven useful.
\citet{huang2018gpipe} designed and trained a 557 million parameter model, AmoebaNet, which achieved 84.4\% top-1 ImageNet accuracy.
Typically, ResNet architectures are scaled up by adding layers (depth): ResNets, suffixed by the number of layers, have marched onward from ResNet-18 to ResNet-200, and beyond~\cite{he2016identity,zhang2020resnest,bello2021lambdanetworks}.
Wide ResNets~\cite{zagoruyko2016wide} and MobileNets~\cite{howard2017mobilenets} instead scale the width.
Increasing image resolutions has also been a reliable source of progress.
Thus as training budgets have grown, so have the image resolutions:  EfficientNet uses 600 image resolutions~\cite{tan2019efficientnet} and both ResNeSt~\cite{zhang2020resnest} and TResNet~\cite{ridnik2020tresnet} use 448 image resolutions for their largest model. 
In an attempt to systematize these heuristics, EfficientNet proposed the compound scaling rule, which recommended balancing the network depth, width and image resolution.
However, Section~\ref{sec:efficientnet_case_study} shows this scaling strategy is sub-optimal for not only ResNets, but EfficientNets as well. 

\paragraph{Additional Training Data.}
Another popular way to further improve accuracy is by training on additional sources of data (either labeled, weakly labeled, or unlabeled).
Pre-training on large-scale datasets~\cite{sun2017revisiting,mahajan2018exploring,kolesnikov2019big} has significantly pushed the state-of-the-art, with ViT~\cite{dosovitskiy2020image} and NFNets~\cite{brock2021highperformance} recently achieving 88.6\% and 89.2\% ImageNet accuracy respectively.
Noisy Student, a semi-supervised learning method, obtained 88.4\% ImageNet top-1 accuracy by using pseudo-labels on an extra 130M unlabeled images~\cite{xie2020self}.
Meta pseudo-labels~\cite{pham2020meta}, an improved semi-supervised learning technique, currently holds the ImageNet state-of-the-art (90.2\%).
We present semi-supervised learning results in Table~\ref{tab:ssl_resnets} and discuss how our training and scaling strategies transfer to large data regimes in Section~\ref{sec:discussion}.

\section{Related Work on Improving ResNets}
Improved training methods combined with architectural changes to ResNets have routinely yielded competitive ImageNet performance~\cite{he2019bag,lee2020compounding,ridnik2020tresnet,zhang2020resnest,bello2021lambdanetworks,brock2021highperformance}.
\citet{he2019bag} achieved 79.2\% top-1 ImageNet accuracy (a +3\% improvement over their ResNet-50 baseline) by modifying the stem and downsampling block while also using label smoothing and mixup. 
\citet{lee2020compounding} further improved the ResNet-50 model with additional architectural modifications such as Squeeze-and-Excitation~\cite{hu2018squeeze}, selective kernel~\cite{li2019selective}, and anti-alias downsampling~\cite{zhang2019making}, while also using label smoothing, mixup, and dropblock to achieve 81.4\% accuracy.
~\citet{ridnik2020tresnet} incorporated several architectural modifications to the ResNet architectures along with improved training methodologies to outperform EfficientNet-B1 to EfficientNet-B5 models on the speed-accuracy Pareto curve.

Most works, however, put little emphasis on identifying strong scaling strategies.
In contrast, we only consider lightweight architectural changes routinely used since 2018 and instead focus on the training and scaling strategies to build a Pareto curve of models.
Our improved training and scaling methods lead to ResNets that are \textbf{1.7x - 2.7x} faster than EfficientNets on TPUs.
Our scaling improvements are orthogonal to the aforementioned methods and we expect them to be additive.

\section{Methodology}
We describe the base ResNet architecture and the training methods used throughout this paper.

\subsection{Architecture}
Our work studies the ResNet architecture, with two widely used architecture changes, the ResNet-D~\cite{he2019bag} modification and Squeeze-and-Excitation (SE) in all bottleneck blocks~\cite{hu2018squeeze}. These architectural changes are used in used many architectures, including TResNet, ResNeSt and EfficientNets.

\textbf{ResNet-D}~\cite{he2019bag} combines the following four adjustments to the original ResNet architecture.
First, the 7$\times$7 convolution in the stem is replaced by three smaller 3$\times$3 convolutions, as first proposed in Inception-V3~\cite{szegedy2016rethinking}.
Second, the stride sizes are switched for the first two convolutions in the residual path of the downsampling blocks.
Third, the stride-2 1$\times$1 convolution in the skip connection path of the downsampling blocks is replaced by stride-2 2$\times$2 average pooling and then a non-strided 1$\times$1 convolution.
Fourth, the stride-2 3$\times$3 max pool layer is removed and the downsampling occurs in the first 3$\times$3 convolution in the next bottleneck block. 
We diagram these modifications in Figure~\ref{fig:resnet_arch_diagram}.

\textbf{Squeeze-and-Excitation}~\cite{hu2018squeeze} reweighs channels via cross-channel interactions by average pooling signals from the entire feature map. 
For all experiments we use a Squeeze-and-Excitation ratio of 0.25 based on preliminary experiments.
In our experiments, we sometimes use the original ResNet implementation without SE (referred to as ResNet) to compare different training methods. 
Clear denotations are made in table captions when this is the case.

\subsection{Training Methods}
We study regularization and data augmentation methods that are routinely used in state-of-the art classification models and semi/self-supervised learning.

\interfootnotelinepenalty=10000
\textbf{Matching the EfficientNet Setup.}
Our training method closely matches that of EfficientNet, where we train for 350 epochs, but with a few small differences. 
\textbf{(1)} We use the cosine learning rate schedule~\cite{loshchilov2016sgdr} instead of an exponential decay for simplicity (no additional hyperparameters). 
\textbf{(2)} We use RandAugment~\cite{cubuk2019randaugment} in all models, whereas EfficientNets were originally trained with AutoAugment~\cite{cubuk2018autoaugment}. 
We reran EfficientNets B0-B4 with RandAugment and found it offered no performance improvement and report EfficientNet B5 and B7 with the RandAugment results from~\citet{cubuk2019randaugment}\footnote{This makes our comparison to EfficientNet-B6 more nuanced as the B6 performance most likely could be improved by 0.1-0.3\% top-1 if ran with RandAugment (based on improvements obtained from B5 and B7).}.
\textbf{(3)} We use the Momentum optimizer instead of RMSProp for simplicity.
See Table~\ref{tab:hparam_comparison} in the Appendix~\ref{sec:appendix_training_details} for a comparison between our training setup and EfficientNet.

\textbf{Regularization.}
We apply \emph{weight decay}, \emph{label smoothing}, \emph{dropout} and \emph{stochastic depth} for regularization. Dropout \cite{srivastava2014dropout} is a common technique used in computer vision and we apply it to the output after the global average pooling occurs in the final layer. 
Stochastic depth ~\cite{huang2016deep} drops out each layer in the network (that has residual connections around it) with a specified probability that is a function of the layer depth. 

\textbf{Data Augmentation.}
We use RandAugment~\cite{cubuk2019randaugment} data augmentation as an additional regularizer. 
RandAugment applies a sequence of random image transformations (e.g. translate, shear, color distortions) to each image independently during training.
As mentioned earlier, originally EfficientNets uses AutoAugment~\cite{cubuk2018autoaugment}, which is a learned augmentation procedure that slightly underperforms RandAugment.

\textbf{Hyperparameter Tuning.}
To select the hyperparameters for the various regularization and training methods, we use a held-out validation set comprising 2\% of the ImageNet training set (20 shards out of 1024).
This is referred to as the \texttt{minival-set} and the original ImageNet validation set (the one reported in most prior works) is referred to as \texttt{validation-set}. The hyperparameters of all ResNet-RS models are in Table~\ref{tab:pareto_curve_hparams} in the Appendix~\ref{sec:pareto_curve_details}.
\section{Improved Training Methods\label{sec:training_setups}}

\subsection{Additive Study of Improvements}
We present an additive study of training, regularization methods and architectural changes in Table~\ref{tab:resnet_method_ablation}.
The baseline ResNet-200 gets 79.0\% top-1 accuracy.
We improve its performance to 82.2\% (\textcolor{blue}{+3.2\%}) through \emph{improved training methods alone} without any architectural changes.
When adding two common and simple architectural changes (Squeeze-and-Excitation and ResNet-D) we further boost the performance to 83.4\%. 
Training methods alone cause $3 / 4$ of the total improvement, which demonstrates their critical impact on ImageNet performance.

\newcommand{\improvement}[1]{\textcolor{blue}{#1}}
\newcommand{\decrease}[1]{\textcolor{red}{#1}}
\newcommand{\improvementb}[1]{\textbf{#1}}
\newcommand{\decreaseb}[1]{#1}

\begin{table}[ht!]
\begin{center}
\small
\begin{tabular}{l|cc}
  \toprule
  Improvements & Top-1 & $\Delta$  \\
  \hline
  ResNet-200 & 79.0 & --- \\
  \rowcolor{blue!15}
  + Cosine LR Decay & 79.3 & \improvementb{+0.3} \\
  \rowcolor{blue!15}
  + Increase training epochs & \;\;78.8 $^{\boldsymbol\dag}$ & \decreaseb{-0.5} \\
  \rowcolor{green!20}
  + EMA of weights & 79.1 & \improvementb{+0.3}\\
  \rowcolor{green!20}
  + Label Smoothing & 80.4 & \improvementb{+1.3} \\
  \rowcolor{green!20}
  + Stochastic Depth & 80.6 & \improvementb{+0.2} \\
  \rowcolor{green!20}
  + RandAugment & 81.0 & \improvementb{+0.4} \\
  \rowcolor{green!20}
  + Dropout on FC & \;\;80.7 $^{\boldsymbol\ddagger}$ & \decreaseb{-0.3}\\
  \rowcolor{green!20}
  + Decrease weight decay & 82.2 & \improvementb{+1.5} \\
  \rowcolor{yellow!20}
  + Squeeze-and-Excitation & 82.9 & \improvementb{+0.7} \\
  \rowcolor{yellow!20}
  + ResNet-D & 83.4 & \improvementb{+0.5}  \\
  \bottomrule
\end{tabular}
\end{center}
\vspace{-0.15cm}
\caption{\textbf{Additive study of the ResNet-RS training recipe.} The colors refer to \textbf{\colorbox{blue!15}{Training Methods}}, \textbf{\colorbox{green!20}{Regularization Methods}} and \textbf{\colorbox{yellow!20}{Architecture Improvements}}. The baseline ResNet-200 was trained for the standard 90 epochs using a stepwise learning rate decay schedule. The image resolution is 256$\times$256.  All numbers are reported on the ImageNet \texttt{validation-set} and averaged over 2 runs. $^{\boldsymbol\dag}$ Increasing training duration to 350 epochs only becomes useful once the regularization methods are used, otherwise the accuracy drops due to over-fitting. $^{\boldsymbol\ddagger}$ dropout hurts as we have not yet decreased the weight decay (See Table~\ref{tab:wd_analysis} for more details).
}
\label{tab:resnet_method_ablation}
\end{table}

\begin{table}[ht!]
\begin{center}
\small
\begin{tabular}{l|l|cc|c}
  \toprule
  Model & Regularization & \multicolumn{2}{c|}{Weight Decay} & $\Delta$ \\
  & & \footnotesize{1e-4} & \footnotesize{4e-5} & \\
  \midrule
  ResNet-50 & None & 79.7 & 78.7 & \textcolor{red}{-1.0} \\
  ResNet-50 & RA-LS & 82.4 & 82.3 & \textcolor{red}{-0.1} \\
  ResNet-50 & RA-LS-DO & 82.2 & 82.7 & \textcolor{blue}{+0.5} \\
  \midrule
  ResNet-200 & None & 82.5 & 81.7 & \textcolor{red}{-0.8} \\
  ResNet-200 & RA-LS & 85.2 & 84.9 & \textcolor{red}{-0.3} \\
  ResNet-200 & RA-LS-SD-DO & 85.3 & 85.5 & \textcolor{blue}{+0.2} \\
  \bottomrule
\end{tabular}
\end{center}
\vspace{-0.15cm}
\caption{\textbf{Decrease weight decay when using more regularization.} Top-1 ImageNet accuracy for different regularization combinations. Decreasing the weight decay improves performance when combining regularization methods such as dropout (DO), stochastic depth (SD), label smoothing (LS) and RandAugment (RA). Image resolution is 224$\times$224 for ResNet-50 and 256$\times$256 for ResNet-200. All numbers are reported on the ImageNet \texttt{minival-set} from an average of two runs.}
\label{tab:wd_analysis} 
\vspace{-0.15cm}
\end{table}

\subsection{Importance of decreasing weight decay when combining regularization methods}
Table~\ref{tab:wd_analysis} highlights the importance of changing weight decay when combining regularization methods together. 
When applying RandAugment and label smoothing, there is no need to change the default weight decay of 1e-4. 
But when we further add dropout and/or stochastic depth, the performance can decrease unless we further decrease the weight decay. 
The intuition is that since weight decay acts as a regularizer, its value must be decreased in order to not overly regularize the model when combining many techniques. 
Furthermore,~\citet{zoph2020learning} presents evidence that the addition of data augmentation shrinks the L2 norm of the weights, which renders some of the effects of weight decay redundant.
Other works use smaller weight decay values, but do not point out the significance of the effect when using more regularization~\cite{tan2019mnasnet,tan2019efficientnet}.

\section{Improved Scaling Strategies \label{sec:scaling_strategies}}
\begin{figure}[t!]
    \begin{center}
    \includegraphics[width=\linewidth]{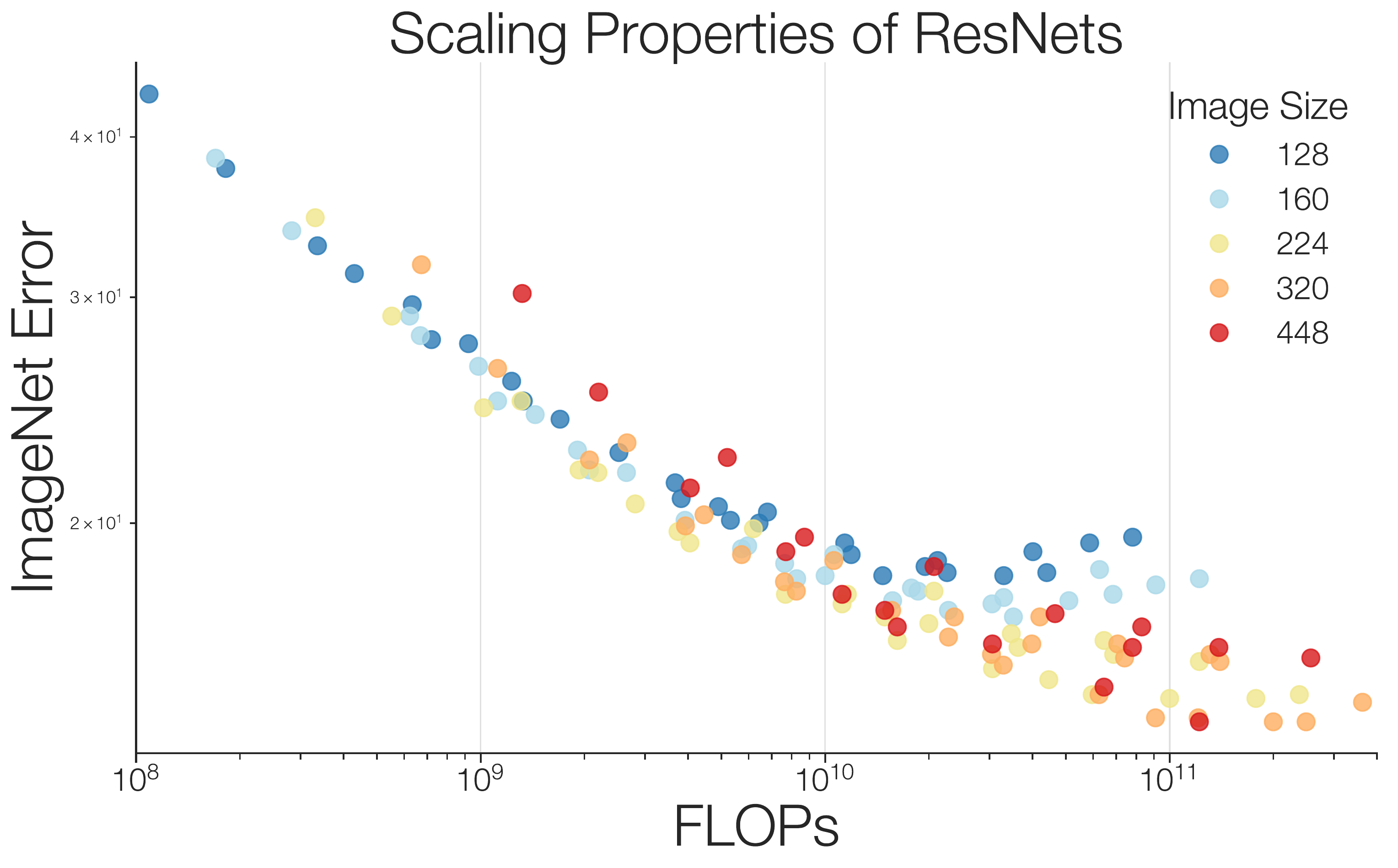}
    \end{center}
    \vspace{-0.3cm}
    \caption{\textbf{Scaling properties of ResNets across varying model scales.} 
    Error approximately scales as a power law with FLOPs (linear fit on the log-log curve) in the lower FLOPs regime but the trend breaks for larger FLOPs.
    We observe diminishing returns of scaling the image resolutions beyond 320$\times$320, which motivates the slow image resolution scaling (Strategy \#2). 
    The scaling configurations run are width multipliers \texttt{[0.25,0.5,1.0,1.5,2.0]}, depths \texttt{[26,50,101,200,300,350,400]} and image resolutions \texttt{[128,160,224,320,448]}.
    FLOPs is the number of floating point operations per image. 
    All results are on the ImageNet \texttt{minival-set}.}
    \label{fig:scaling_image_size}
    \vspace{-0.1cm}
\end{figure}

The prior section demonstrates the significant impact of training methodology and we now show the scaling strategy is similarly important.
In order to establish scaling trends, we perform an extensive search on ImageNet over width multipliers in \texttt{[0.25,0.5,1.0,1.5,2.0]}, depths of \texttt{[26,50,101,200,300,350,400]} and resolutions of \texttt{[128,160,224,320,448]}.
We train these architectures for 350 epochs, mimicking the training setup of state-of-the-art ImageNet models.
We increase the regularization as the model size increases to limit overfitting. 
See Appendix~\ref{sec:reg_scheme_scaling} for regularization and model hyperparameters.

\begin{figure*}[ht!]
    \begin{center}
    \includegraphics[width=0.33\textwidth]{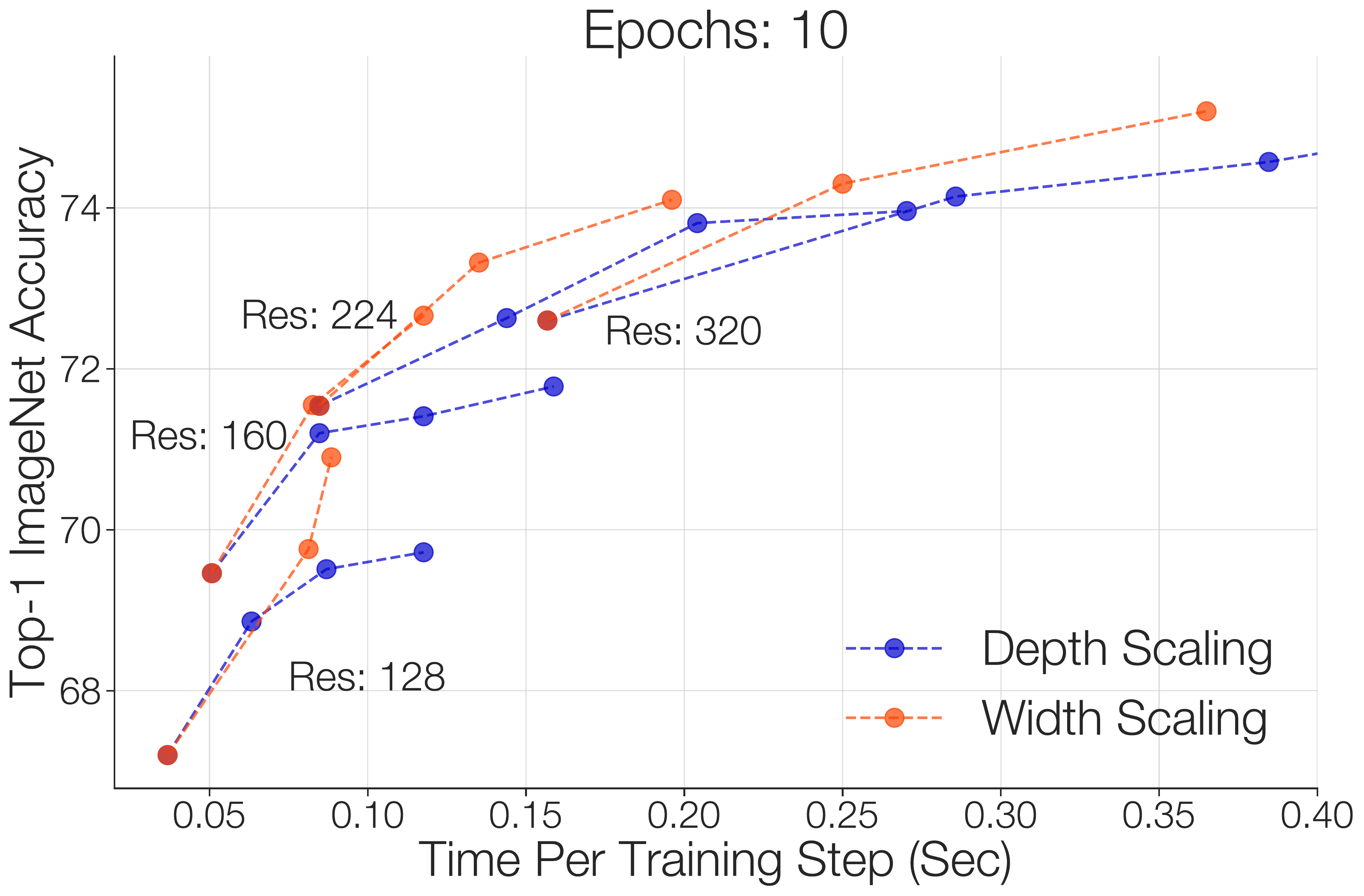}
    \includegraphics[width=0.33\textwidth]{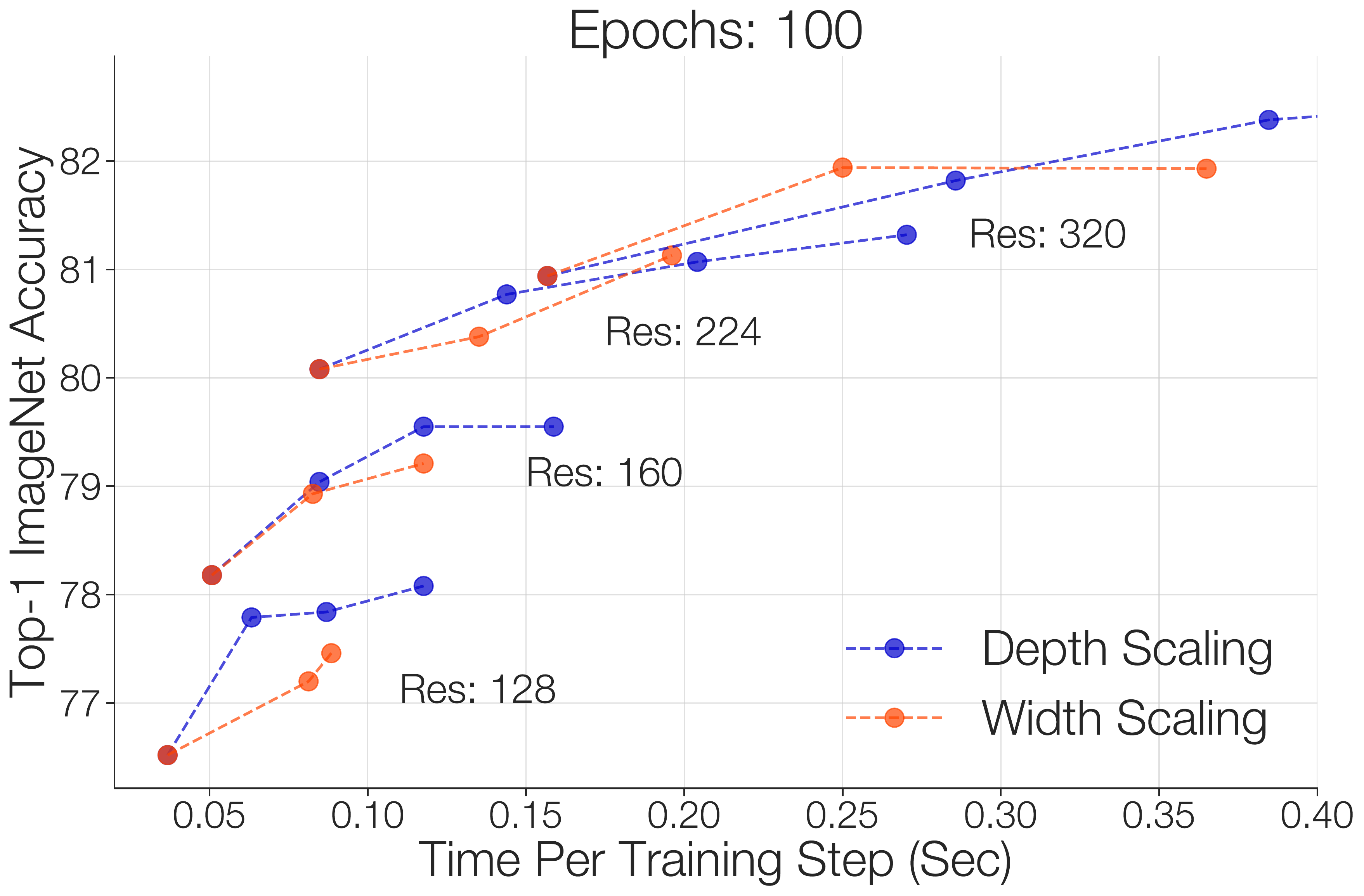}
    \includegraphics[width=0.33\textwidth]{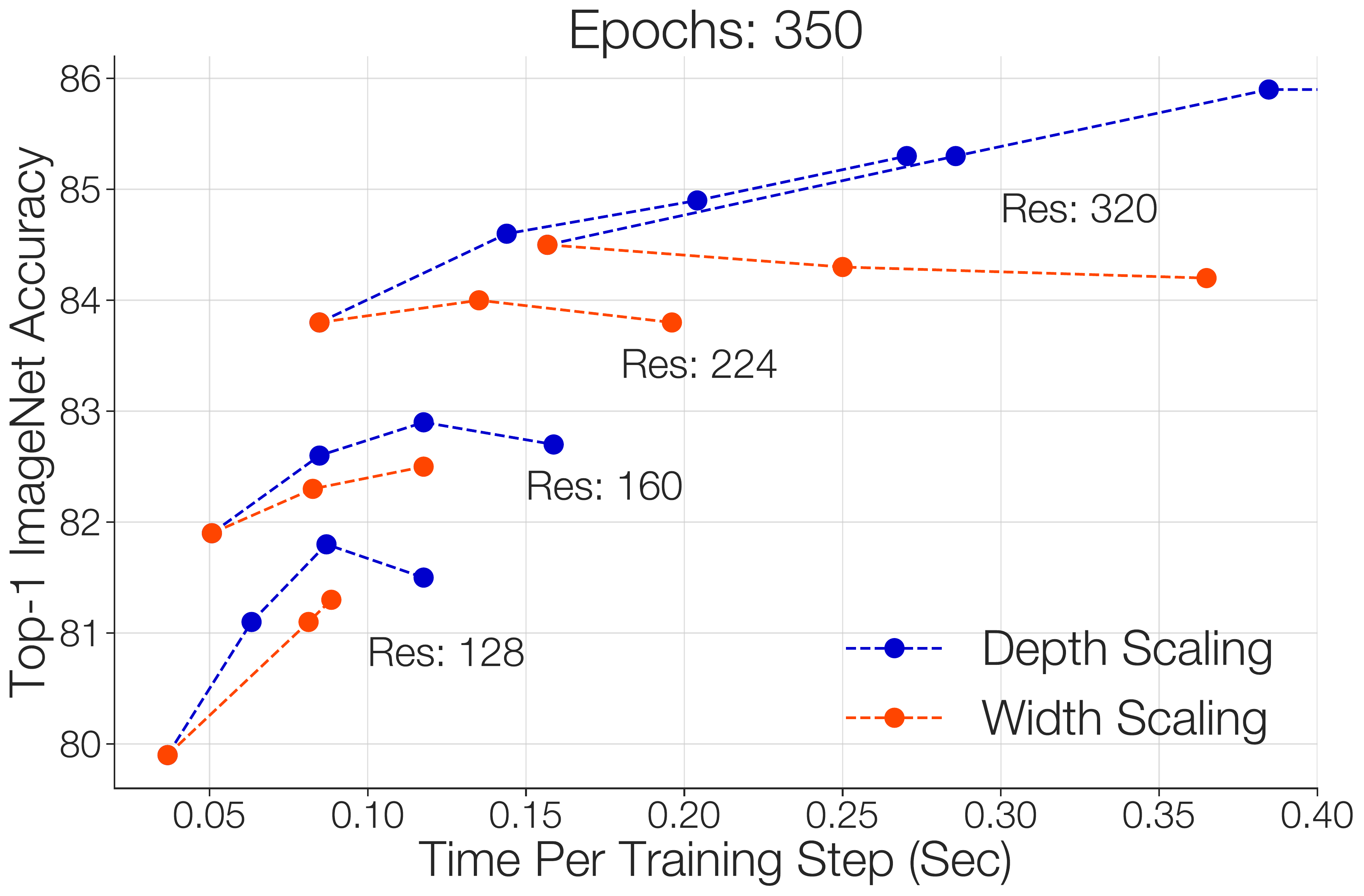}
    \end{center}
    \vspace{-0.3cm}
    \caption{\textbf{Scaling of ResNets across depth, width, image resolution and training epochs}. We compare depth scaling and width scaling across four different image resolutions \texttt{[128,160,224,320]} when training models for 10, 100 or 350 epochs.
    We find that \emph{the best performing scaling strategy depends on the training regime}, which reveals the pitfall of extrapolating scaling rules from small scale regimes.
    \textbf{(Left) 10~Epoch~Regime}: width scaling is the best strategy for the speed-accuracy Pareto curve.
    \textbf{(Middle) 100~Epoch~Regime}: depth scaling is sometimes outperformed by width scaling.
    \textbf{(Right) 350~Epoch~Regime}: depth scaling consistently outperforms width scaling by a large margin. Overfitting remains an issue even when using regularization methods.
    \textbf{Model Details:} All models start from a depth of 101 and are increased through \texttt{[101,200,300,400]}. All model widths start with a multiplier of \texttt{1.0x} and are increased through \texttt{[1.0,1.5,2.0]}.
    For all models, we tune regularization in an effort to limit overfitting (see Appendix~\ref{sec:reg_scheme_scaling}).
    Accuracies are reported on the ImageNet \texttt{minival-set} and training times are measured on TPUs.}
    \label{fig:scaling_analysis}
    \vspace{-0.1cm}
\end{figure*}

\paragraph{FLOPs do not accurately predict performance in the bounded data regime.}
Prior works on scaling laws observe a power law between error and FLOPs in \emph{unbounded data regimes}~\cite{kaplan2020scaling,henighan2020scaling}.
In order to test whether this also holds in our scenario, we plot ImageNet error against FLOPs for all scaling configurations in Figure~\ref{fig:scaling_image_size}.
For the smaller models, we observe an overall power law trend between error and FLOPs, with minor dependency on the scaling configuration (i.e.\ depth versus width versus image resolution).
However, the trend breaks for larger model sizes. 
Furthermore, we observe a large variation in ImageNet performance for a fixed amount of FLOPs, especially in the higher FLOP regime. 
Therefore the exact scaling configuration (i.e.\ depth, width and image resolution) can have a big impact on performance even when controlling for the same amount of FLOPs.

\paragraph{The best performing scaling strategy depends on the training regime.}
We next look directly at latencies\footnote{FLOPs is not a good indicator of latency on modern hardware. See Section~\ref{sec:resnet_rs_speed} for a more detailed discussion.} on the hardware of interest to identify scaling strategies that improve the speed-accuracy Pareto curve.
Figure~\ref{fig:scaling_analysis} presents accuracies and latencies of models scaled with either width or depth across four image resolutions and three different training regimes (10, 100 and 350 epochs).
We observe that the best performing scaling strategy, especially whether to scale depth and/or width, highly depends on the training regime.

\subsection{Strategy \#1 - Depth Scaling in Regimes Where Overfitting Can Occur}
\paragraph{Depth scaling outperforms width scaling for longer epoch regimes.} 
In the 350 epochs setup (Figure~\ref{fig:scaling_analysis}, right panel), we observe depth scaling to significantly outperform width scaling across all image resolutions.
Scaling the width is subject to overfitting and sometimes hurts performance even with increased regularization.
We hypothesize that this is due to the larger increase in parameters when scaling the width. 
The ResNet architecture maintains constant FLOPs across all block groups and multiplies the number of parameters by 4$\times$ every block group.
Scaling the depth, especially in the earlier layers, therefore introduces fewer parameters compared to scaling the width.

\paragraph{Width scaling outperforms depth scaling for shorter epoch regimes.} 
In contrast, width scaling is better when only training for 10 epochs ( Figure~\ref{fig:scaling_analysis}, left panel).
For 100 epochs (Figure~\ref{fig:scaling_analysis}, middel panel), the best performing scaling strategy varies between depth scaling and width scaling, depending on the image resolution.
The dependency of the scaling strategy on the training regime reveals a pitfall of extrapolating scaling rules.
We point out that prior works also choose to scale the width when training for a small number of epochs on large-scale datasets (e.g.\ $\sim$40 epochs on 300M images), consistent with our experimental findings that scaling the width is preferable in shorter epoch regimes.
In particular, ~\citet{kolesnikov2019big} train a ResNet-152 with 4x filter multiplier while \citet{brock2021highperformance} scales the width with $\sim$1.5x filter multiplier.

\subsection{Strategy \#2 - Slow Image Resolution Scaling}
In Figure~\ref{fig:scaling_image_size}, we also observe that larger image resolutions yield diminishing returns.
We therefore propose to increase the image resolution more gradually than previous works.
This contrasts with the compound scaling rule proposed by EfficientNet which leads to very large images (e.g. 600 for EfficientNet-B7, 800 for EfficientNet-L2~\citep{xie2020self}).
Other works such as ResNeSt~\cite{zhang2020resnest} and TResNet~\cite{ridnik2020tresnet}) scale the image resolution up to 448.
Our experiments indicate that slower image scaling improves not only ResNet architectures, but also EfficientNets on a speed-accuracy basis (Section~\ref{sec:efficientnet_case_study}).

\subsection{Two Common Pitfalls in Designing Scaling Strategies} 
Our scaling analysis surfaces two common pitfalls in prior research on scaling strategies: 

\textbf{(1) Extrapolating scaling strategies from small-scale regimes.}
Scaling strategies found in small scale regimes (e.g. on small models or with few training epochs) can fail to generalize to larger models or longer training iterations.
The dependencies between the best performing scaling strategy and the training regime are missed by prior works which extrapolate scaling rules from either small models~\cite{tan2019efficientnet} or shorter training epochs~\cite{radosavovic2020designing}.
We therefore do not recommend generating scaling rules exclusively in a small scale regime because these rules can break down.

\textbf{(2) Extrapolating scaling strategies from a single and potentially sub-optimal initial architecture.}
Beginning from a sub-optimal initial architecture can skew the scaling results.
For example, the compound scaling rule derived from a small grid search around EfficientNet-B0, which was obtained by architecture search using a fixed FLOPs budget and a specific \emph{image resolution}.
However, since this image resolution can be sub-optimal for that FLOPs budget, the resulting scaling strategy can be sub-optimal.
In contrast, our work designs scaling strategies by training models across a variety of widths, depths and image resolutions.

\subsection{Summary of Improved Scaling Strategies}
For a new task, we recommend running a \emph{small subset} of models across different scales, for the full training epochs, to gain intuition on which dimensions are the most useful across model scales. 
While this approach may appear more costly, we point out that the cost is offset by not searching for the architecture.

For image classification, the scaling strategies are summarized as \textbf{(1)} scale the depth in regimes where overfitting can occur (scaling the width is preferable otherwise) and \textbf{(2)} slow image resolution scaling.
Experiments indicate that applying these scaling strategies to ResNets (ResNet-RS) and EfficientNets (EfficientNet-RS) leads to significant speed-ups over EfficientNets. 
We note that similar scaling strategies are also employed in recent works that obtain large speed-ups over EfficientNets such as LambdaResNets~\cite{bello2021lambdanetworks} and NFNets~\cite{brock2021highperformance}.
\section{Experiments with Improved Training and Scaling Strategies}
\begin{figure}[t!]
    \begin{center}
    \includegraphics[width=\linewidth]{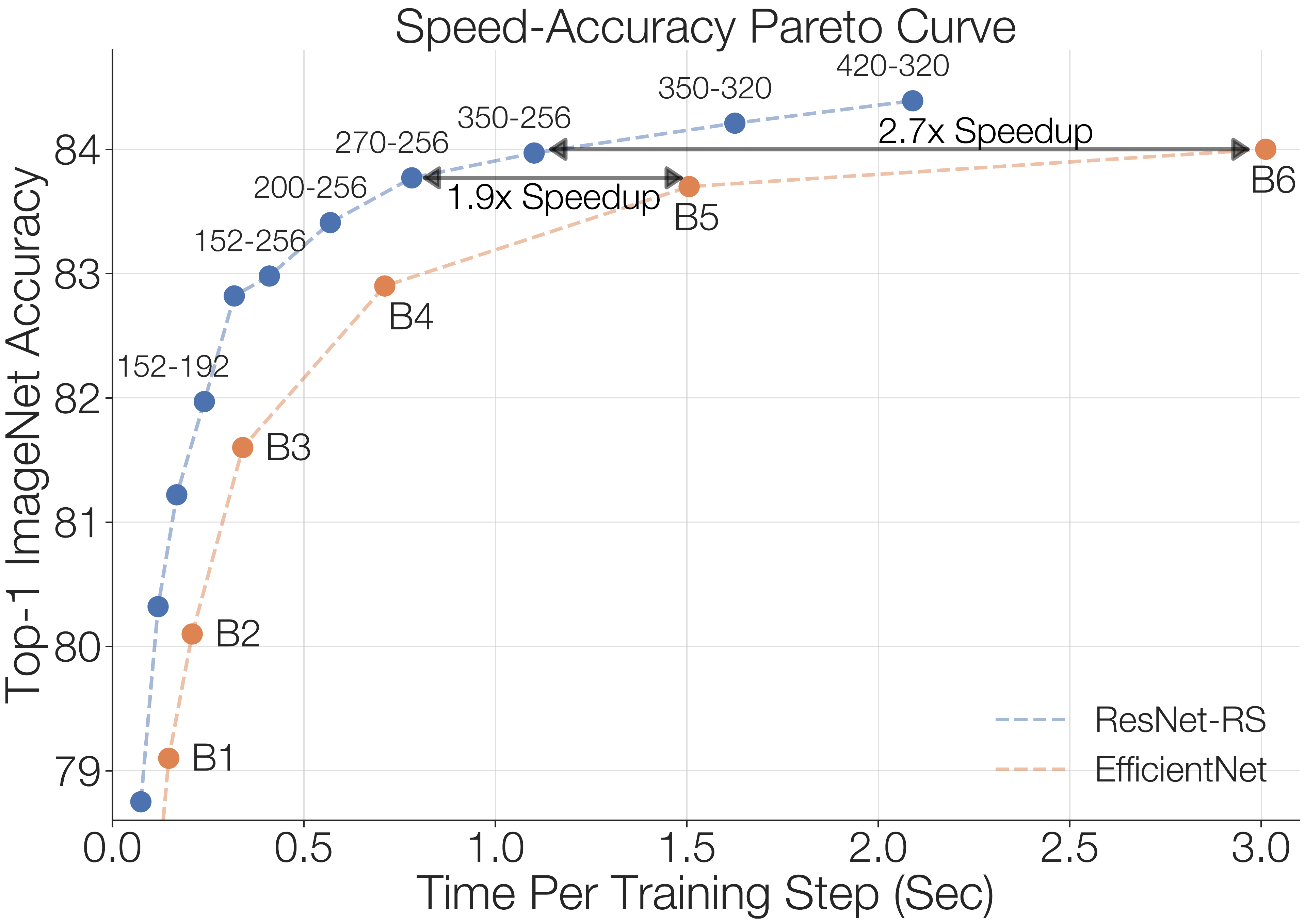}
    \end{center}
    \vspace{-0.4cm}
    \caption{\textbf{Speed-Accuracy Pareto curve comparing ResNets-RS to EfficientNet.} 
    Properly scaled ResNets (ResNet-RS) are \textbf{1.7x - 2.7x} faster than the popular EfficientNets when closely matching their training setup. 
    ResNet-RS are annotated with (depth - image resolution), so 152-256 means ResNet-RS-152 with image resolution 256$\times$256. 
    All results are on the ImageNet \texttt{validation-set} and training times are measured on TPUs.
    See Appendix~\ref{sec:pareto_curve_details} for detailed results.}
    \label{fig:pareto_curve_inference}
    \vspace{-0.1cm}
\end{figure}

\subsection{ResNet-RS on a Speed-Accuracy Basis\label{sec:resnet_rs_speed}}
Using the improved training and scaling strategies, we design \emph{ResNet-RS}, a family of re-scaled ResNets across a wide range of model scales (see Appendix~\ref{sec:pareto_curve_details}~and~\ref{sec:architectural_details} for experimental and architectural details).
Figure~\ref{fig:pareto_curve_inference} compares EfficientNets against ResNet-RS on a speed-accuracy Pareto curve.
We find that ResNet-RS match EfficientNets' performance while being \emph{1.7x - 2.7x} faster on TPUs.
This large speed-up over EfficientNet may be non-intuitive since EfficientNets significantly reduce both the parameter count and the FLOPs compared to ResNets.
We next discuss why a model with fewer parameters and fewer FLOPs (EfficientNet) is slower and more memory-intensive during training.

\paragraph{FLOPs vs Latency.}
While FLOPs provide a hardware-agnostic metric for assessing computational demand, they may not be indicative of actual latency times for training and inference~\cite{howard2017mobilenets,howard2019searching,radosavovic2020designing}.
In custom hardware architectures (e.g. TPUs and GPUs), FLOPs are an especially poor proxy because operations are often bounded by memory access costs and have different levels of optimization on modern matrix multiplication units~\cite{jouppi2017datacenter}.
The inverted bottlenecks~\cite{sandler2018mobilenetv2} used in EfficientNets employ depthwise convolutions with large activations and have a \emph{small compute to memory ratio} (operational intensity) compared to the ResNet's bottleneck blocks which employ dense convolutions on smaller activations.
This makes EfficientNets less efficient on modern accelerators compared to ResNets.
Table~\ref{tab:mem_and_flops_comp} illustrates this point: a ResNet-RS model with \textbf{1.8x} more FLOPs than EfficientNet-B6 is \textbf{2.7x} faster on a TPUv3 hardware accelerator.

\paragraph{Parameters vs Memory.}
Parameter count does not necessarily dictate memory consumption during \emph{training} because memory is often dominated by the size of the \emph{activations}\footnote{Activations are typically stored during training as they are used in backpropagation. At inference, activations can be discarded and parameter count is a better proxy for actual memory consumption.}.
The large activations used in EfficientNets also cause larger memory consumption, which is exacerbated by the use of large image resolutions, compared to our re-scaled ResNets.
A ResNet-RS model with \textbf{3.8x} more parameters than EfficientNet-B6 consumes \textbf{2.3x} less memory for a similar ImageNet accuracy (Table~\ref{tab:mem_and_flops_comp}).
We emphasize that both memory consumption and latency are tightly coupled to the software and hardware stack (TensorFlow on TPUv3) due to compiler optimizations such as operation layout assignments and memory padding.

\begin{table}[t!]
\begin{center}
\small
\begin{tabular}{l|cc|cc}
  \toprule
  \rowcolor{gray!15}
  \textbf{Model} & \textbf{RS-350} & \textbf{ENet-B6} & \textbf{RS-420} & \textbf{ENet-B7} \\
  \midrule
  Resolution & 256 & 528 & 320 & 600 \\
  \midrule
  \rowcolor{gray!15}
  Top-1 Acc. & \textbf{84.0} & \textbf{84.0} & 84.4 & \textbf{84.7} \\
  \midrule
  Params \textcolor{gray}{(M)} & 164 & 43\;\textcolor{blue}{\scriptsize{(3.8x)}} & 192 & 66\;\textcolor{blue}{\scriptsize{(2.9x)}} \\
  FLOPs \textcolor{gray}{(B)} & 69 & 38\;\textcolor{blue}{\scriptsize{(1.8x)}} & 128 & 74\;\textcolor{blue}{\scriptsize{(1.7x)}} \\
  \midrule
  \rowcolor{gray!15}
  TPU-v3 & & & & \\
  \;\;Latency \textcolor{gray}{(s)} & 1.1 & 3.0\;\textcolor{red}{\scriptsize{(2.7x)}} & 2.1 & 6.0\;\textcolor{red}{\scriptsize{(2.9x)}} \\
  \;\;Memory \textcolor{gray}{(GB)} & 7.3 & 16.6\;\textcolor{red}{\scriptsize{(2.3x)}} & 15.5 & 28.3\;\textcolor{red}{\scriptsize{(1.8x)}} \\
  \midrule
  \rowcolor{gray!15}
  V100  & & & & \\
  \;\;Latency \textcolor{gray}{(s)} & 4.7 & 15.7\;\textcolor{red}{\scriptsize{(3.3x)}} & 10.2 & 29.9\;\textcolor{red}{\scriptsize{(2.8x)}} \\
  \bottomrule
\end{tabular}
\end{center}
\vspace{-0.2cm}
\caption{\textbf{Performance comparison of ResNet-RS and EfficientNet} (abbreviated ENet). Although ResNet-RS has more parameters and FLOPs, the model employs less memory and runs faster on TPUs and GPUs. TPU latency is reported as the time per training step for 1024 images on 8 TPUv3 cores. Memory is reported on 32 images per core, using \texttt{bfloat16} precision without fusion or rematerialization. See Appendix~\ref{sec:profiling} for more profiling details.}
\label{tab:mem_and_flops_comp}
\end{table}

\subsection{Improving the Efficiency of EfficientNets \label{sec:efficientnet_case_study}}
\begin{figure}[t!]
    \begin{center}
    \includegraphics[width=\linewidth]{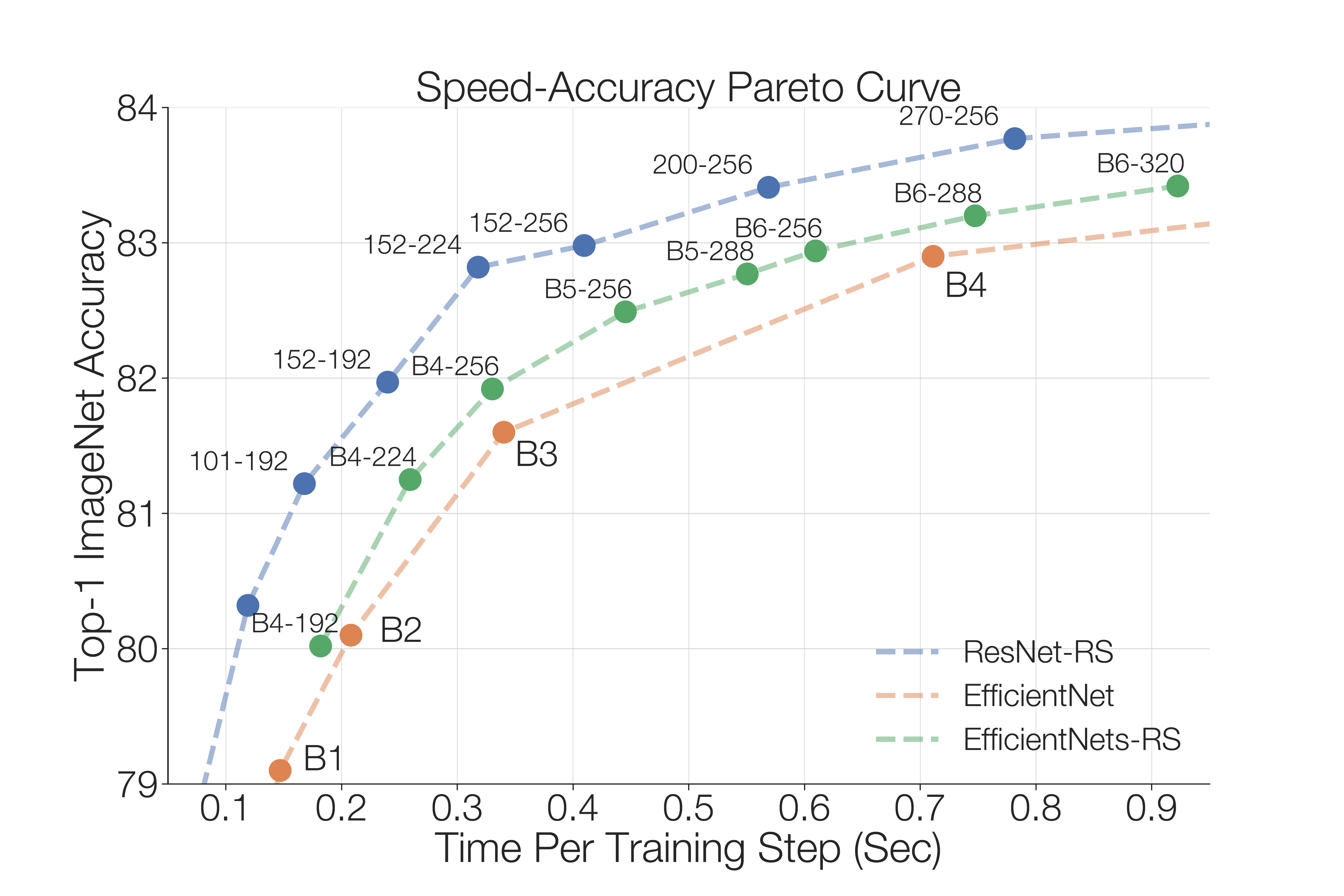}
    \end{center}
    \vspace{-0.4cm}
    \caption{\textbf{Speed-Accuracy Pareto curve comparing ResNets-RS and EfficientNet-RS to EfficientNet.} Scaling EfficientNets using the slow image resolution scaling strategy (instead of the original compound scaling rule) improves the Pareto efficiency of EfficientNets. Note that ResNet-RS still outperforms EfficientNet-RS. This figure is a zoomed in version of Figure~\ref{fig:pareto_curve_inference} with EfficientNet-RS added. 
    Models are annotated with (model depth - image resolution), so 152-192 corresponds to ResNet-RS-152 with image resolution 192$\times$192. 
    Results are reported on the ImageNet \texttt{validation-set} and training times are measured on TPUs.}
    \label{fig:enet_rs_pareto_curve_inference}
    \vspace{-0.1cm}
\end{figure}

The scaling analysis from Section~\ref{sec:scaling_strategies} reveals that scaling the image resolution results in \emph{diminishing} returns. 
This suggests that the scaling rules advocated in EfficientNets which increases model depth, width and resolution \emph{independently} of model scale is sub-optimal. 
We apply the slow image resolution scaling strategy (Strategy \#2) to EfficientNets and train several versions with reduced image resolutions, without changing the width or depth. 
The RandAugment magnitude is set to 10 for image resolution 224 or smaller, 20 for image resolution larger than 320 and 15 otherwise.
All other hyperparameters are kept the same as per the original EfficientNets.
Figure~\ref{fig:enet_rs_pareto_curve_inference} demonstrates a marked improvement of the re-scaled EfficientNets (EfficientNet-RS) on the speed-accuracy Pareto curve over the original EfficientNets.

\subsection{Semi-Supervised Learning with ResNet-RS}
We measure how ResNet-RS performs as we scale to larger datasets in a large scale semi-supervised learning setup. 
We train ResNets-RS on the combination of 1.2M labeled ImageNet images and 130M \emph{pseudo-labeled} images, in a similar fashion to Noisy Student~\cite{xie2020self}. 
We use the same dataset of 130M images pseudo-labeled as Noisy Student, where the pseudo labels are generated from an EfficientNet-L2 model with 88.4\% ImageNet accuracy.
Models are jointly trained on both the labeled and pseudo-labeled data and training hyperparameters are kept the same.
Table~\ref{tab:ssl_resnets} reveals that ResNet-RS models are very strong in the semi-supervised learning setup as well. We obtain a top-1 ImageNet accuracy of 86.2\%, while being \textbf{4.7x} faster on TPU (\textbf{5.5x} on GPU) than the corresponding Noisy Student EfficientNet-B5 model.
\begin{table}[ht!]
    \centering
    \begin{tabular}{lccc}
      \toprule
      Model & V100 \textcolor{gray}{(s)} & TPUv3 \textcolor{gray}{(ms)} & Top-1 \\
      \midrule
      EfficientNet-B5 & 8.16 & 1510 & 86.1 \\
      ResNet-RS-152 & \textbf{1.48} \rlap{\textbf{\textcolor{blue}{(5.5x)}}} & \textbf{320} \rlap{\textbf{\textcolor{blue}{(4.7x)}}} & \textbf{86.2} \\
      \bottomrule
    \end{tabular}
    \caption{\textbf{ResNet-RS are efficient semi-supervised learners.} 
    ResNet-RS-152 with image resolution 224 is \textbf{4.7x} faster on TPU (\textbf{5.5x} on GPU) than EfficientNet-B5 Noisy Student~\cite{xie2020self} for a similar ImageNet accuracy. 
    Both models train on the same additional 130M pseudo-labeled images. 
    See Appendix~\ref{sec:profiling} for details on latency measurements.}
    \label{tab:ssl_resnets} 
    \vspace{-0.15cm}
\end{table}

\definecolor{myblue}{RGB}{20, 23, 204}
\begin{table*}[h!]
    \centering
    \begin{tabular}{lcc|ccccc}
      \toprule
      Model & Training & Epochs & CIFAR-100 & Pascal & Pascal & ADE & NYU \\
      & Method & &  Accuracy & Detection & Segmentation & Segmentation & Depth   \\
      \midrule
      ResNet-152 & Supervised & 90 & 85.5 & 80.0 & 70.0 & 40.2 & 81.2 \\
      ResNet-152 & SimCLR & 800 & 87.1 & \textcolor{myblue}{\textbf{83.3}} & 72.2 & 41.0 & 83.5 \\
      ResNet-152 & SimCLRv2 & 800 & 84.7 & 79.1 & 73.1 & 41.1 & \textcolor{myblue}{\textbf{84.7}} \\
      ResNet-152 & RS & 400 & \textcolor{myblue}{\textbf{88.1}} & 82.2 & \textcolor{myblue}{\textbf{78.2}} & \textcolor{myblue}{\textbf{42.2}} & 83.4 \\
      \midrule
      ResNet-152 2x & Supervised & 90 & 86.6 & 81.1 & 72.2 & 41.4 & 82.5 \\
      ResNet-152 2x & SimCLR & 800 & 89.0 & \textcolor{myblue}{\textbf{85.3}} & 78.8 & \textcolor{myblue}{\textbf{45.2}} & \textcolor{myblue}{\textbf{86.8}}\\
      ResNet-152 2x & SimCLRv2 & 800 & 84.8 & 80.1 & 75.5 & 42.5 & 86.1 \\
      ResNet-152 2x & RS & 400 & \textcolor{myblue}{\textbf{89.3}} & 84.1 & \textcolor{myblue}{\textbf{79.2}} & 44.1 & 84.7  \\
      \bottomrule
    \end{tabular}
    \caption{\textbf{Representations from supervised learning with improved training strategies rival or outperform representations from state-of-the-art self-supervised learning algorithms.} 
    Comparison of supervised training methods (supervised, RS) and self-supervised methods (SimCLR, SimCLRv2) on a variety of downstream tasks. 
    The (RS) strategy greatly outperforms the baseline supervised training, which highlights the importance of using improved supervised training techniques when comparing to self-supervised learning algorithms.
    The RS training method uses a subset of the training methods highlighted in this work (cosine LR decay, RandAugment label smoothing, reduced weight decay, and dropout on FC) to more closely match those used in the self-supervised algorithms. 
    All models employ the \emph{vanilla} ResNet architecture without modifications and are pre-trained on ImageNet.}
    \label{tab:suite_transfer} 
    \vspace{-0.15cm}
\end{table*}

\subsection{Transfer Learning of ResNet-RS}
We now investigate whether the improved supervised training strategies yield better representations for transfer learning and compare them with self-supervised learning algorithms.
Recent self-supervised learning algorithms claim to surpass the transfer learning performance of \emph{supervised learning} and create more universal representations \cite{chen2020simple,chen2020big}.
Self-supervised algorithms, however, make several changes to the training methods (e.g training for more epochs, data augmentation) making comparisons to supervised learning difficult.
Table~\ref{tab:suite_transfer} compares the transfer performance of improved supervised training strategies (denoted RS) against self-supervised SimCLR~\cite{chen2020simple} and SimCLRv2~\cite{chen2020big}.
In an effort to closely match SimCLR's training setup and provide fair comparisons, we restrict the RS training strategies to a subset of its original methods.
Specifically, we employ data augmentation (RandAugment), label smoothing, dropout, decreased weight decay and cosine learning rate decay for 400 epochs but do not use stochastic depth or exponential moving average (EMA) of the weights.
We choose this subset to closely match the training setup of SimCLR: longer training, data augmentation and a temperature parameter for their contrastive loss\footnote{Note that SimCLR and SimCLRv2 might benefit further when combining with RandAugment, but the same may also hold true when combining SimCLR's augmentation with RandAugment for supervised learning.}.
We use the vanilla ResNet architecture without the ResNet-D modifications or Squeeze-and-Excite, matching the SimCLR and SimCLRv2 architectures.

We evaluate the transfer performance on five downstream tasks: CIFAR-100 Classification~\cite{krizhevsky2009learning}, Pascal Detection \& Segmentation~\cite{everingham2010pascal}, ADE Segmentation~\cite{zhou2017scene} and NYU Depth~\cite{silberman2012indoor}.
We find that, even when restricted to a smaller subset, the improved training strategies improve transfer performance\footnote{\citet{kornblith2019better} similarly observed that better ImageNet top-1 accuracy (either through better architectures or training strategies) strongly correlates with improved transfer learning performance.}. 
The improved supervised representations (RS) outperform SimCLR on \textbf{$5 / 10$} downstream tasks and SimCLRv2 on \textbf{$8 / 10$} tasks. 
Furthermore, the improved training strategies significantly outperform the standard supervised ResNet representations, highlighting the need for using modern training techniques when comparing to self-supervised learning.
While self-supervised learning can be used on \emph{unlabeled data}, our results challenge the notion that self-supervised algorithms lead to more universal representations than supervised learning when labels are available.

\subsection{Revised 3D ResNet for Video Classification}
\begin{table}[t!]
\begin{center}
\small
\begin{tabular}{l|cc}
  \toprule
  Improvements & Top-1 & $\Delta$  \\
  \midrule
  3D ResNet-50 & 73.4 & -- \\
  \rowcolor{green!20}
  + Dropout on FC & 74.4 & \improvementb{+1.0}\\
  \rowcolor{green!20}
  + Label smoothing & 74.9 & \improvementb{+0.5} \\
  \rowcolor{green!20}
  + Stochastic depth & 76.1 & \improvementb{+1.2} \\
  \rowcolor{green!20}
  + EMA of weights & 76.1 & -- \\
  \rowcolor{green!20}
  + Decrease weight decay & 76.3 & \improvementb{+0.2} \\
  \rowcolor{blue!15}
  + Increase training epochs & 76.4 & \improvementb{+0.1} \\
  \rowcolor{green!20}
  + Scale jittering & 77.4 & \improvementb{+1.0} \\
  \rowcolor{yellow!20}
  + Squeeze-and-Excitation & 77.9 & \improvementb{+0.5} \\
  \rowcolor{yellow!20}
  + ResNet-D & 78.2 & \improvementb{+0.3}  \\
  \bottomrule
\end{tabular}
\end{center}
\caption{\textbf{Additive study of training methods for video classification.} The colors refer to \textbf{\colorbox{blue!15}{Training Methods}}, \textbf{\colorbox{green!20}{Regularization Methods}} and \textbf{\colorbox{yellow!20}{Architecture Improvements}}. The ResNet-RS training recipe transfers to a 3D ResNet model on Kinetics-400 video classification \cite{Kay2017TheKH}. Reported accuracies are averaged over 2 runs. The baseline 3D ResNet-50 was trained for 200 epochs with a cosine learning rate decay.}
\label{tab:ResNet3D Performance} 
\vspace{-0.25cm}
\end{table}

We conclude by applying the training strategies to the Kinetics-400 video classification task, using a 3D ResNet as the baseline architecture~\cite{Qian2020SpatiotemporalCV} (see Appendix~\ref{sec:video_classification_details} for experimental details).
Table~\ref{tab:ResNet3D Performance} presents an additive study of the RS training recipe and architectural improvements.

The training strategies extend to video classification, yielding a combined improvement from $73.4\%$ to $77.4\%$ (\improvement{+4.0\%}).
The ResNet-D and Squeeze-and-Excitation architectural changes further improve the performance to $78.2\%$ (\improvement{+0.8\%}).
Similarly to our study on image classification (Table~\ref{tab:resnet_method_ablation}), we find that most of the improvement can be obtained without architectural changes.
Without model scaling, 3D ResNet-RS-50 is only 2.2\% less than the best number reported on Kinetics-400 at 80.4\%~\cite{Feichtenhofer2020X3DEA}.
\section{Discussion\label{sec:discussion}}

\paragraph{Why is it important to tease apart improvements coming from training methods vs architectures?}
Training methods can be more task-specific than architectures (e.g. data augmentation is more helpful on small datasets).
Therefore, improvements coming from training methods do not necessarily generalize as well as architectural improvements.
Packaging newly proposed architectures together with training improvements makes accurate comparisons between architectures difficult.
The large improvements coming from training strategies, when not being controlled for, can overshadow architectural differences.

\vspace{-0.25cm}
\paragraph{How should one compare different architectures?}
Since training methods and scale typically improve performance~\cite{lee2020compounding,kaplan2020scaling}, it is critical to control for both aspects when comparing different architectures.
Controlling for scale can be achieved through different metrics.
While many works report parameters and FLOPs, we argue that latencies and memory consumption are generally more relevant~\cite{radosavovic2020designing}.
Our experimental results (Section~\ref{sec:resnet_rs_speed}) re-emphasize that FLOPs and parameters are not representative of latency or memory consumption~\cite{radosavovic2020designing,norrie2021design}.

\vspace{-0.25cm}
\paragraph{Do the improved training strategies transfer across tasks?}
The answer depends on the domain and dataset sizes available.
Many of the training and regularization methods studied here are not used in large-scale pre-training (e.g. 300M images) \cite{kolesnikov2019big,dosovitskiy2020image}.
Data augmentation is useful for small datasets or when training for many epochs, but the specifics of the augmentation method can be task-dependent (e.g. scale jittering instead of RandAugment in Table~\ref{tab:ResNet3D Performance}).

\vspace{-0.25cm}
\paragraph{Do the scaling strategies transfer across tasks?}
The best performing scaling strategy depends on the training regime and whether overfitting is an issue, as discussed in Section~\ref{sec:scaling_strategies}.
When training for 350 epochs on ImageNet, we find scaling the depth to work well, whereas scaling the width is preferable when training for few epochs (e.g. 10 epochs).
This is consistent with works employing width scaling when training for few epochs on large-scale datasets~\cite{kolesnikov2019big}.
We are unsure how our scaling strategies apply in tasks that require larger image resolutions (e.g. detection and segmentation) and leave this to future work.

\vspace{-0.25cm}
\paragraph{Are architectural changes useful?}
Yes, but training methods and scaling strategies can have even larger impacts.
Simplicity often wins, especially given the non-trivial performance issues arising on custom hardware.
Architecture changes that decrease speed and increase complexity may be surpassed by scaling up faster and simpler architectures that are optimized on available hardware (e.g convolutions instead of depthwise convolutions for GPUs/TPUs).
We envision that future successful architectures will emerge by co-design with hardware, particularly in resource-tight regimes like mobile phones~\cite{howard2019searching}.

\vspace{-0.25cm}
\paragraph{How should one allocate a computational budget to produce the best vision models?}
We recommend beginning with a simple architecture that is efficient on available hardware (e.g. ResNets on GPU/TPU) and training several models, to convergence, with different image resolutions, widths and depths to construct a Pareto curve.
Note that this strategy is distinct from \citet{tan2019efficientnet} which instead allocate a large portion of the compute budget for identifying an optimal initial architecture to scale.
They then do a small grid search to find the compound scaling coefficients used across all model scales. 
RegNet~\cite{radosavovic2020designing} does most of their studies when training for only 10 epochs.

\section{Conclusion}
By updating the de facto vision baseline with modern training methods and an improved scaling strategy, we have revealed the remarkable durability of the ResNet architecture.
Simple architectures set strong baselines for state-of-the-art methods.
We hope our work encourages further scrutiny in maintaining consistent methodology for both proposed innovations and baselines alike.

{\footnotesize \paragraph{Acknowledgements.} We would like to thank Ashish Vaswani, Prajit Ramachandran, Ting Chen, Thang Luong, Hanxiao Liu, Gabriel Bender, Quoc Le, Neil Houlsby, Mingxing Tan, Andrew Howard, Raphael Gontijo Lopes, Andy Brock and David Berthelot for helpful feedback on this work; Jing Li, Pengchong Jin, Yeqing Li and Yin Cui for the support on open-sourcing and infrastructure.}


\bibliography{main}
\bibliographystyle{icml2021}

\clearpage 
\onecolumn
\appendix
\section{Author Contributions}

\textbf{IB,~BZ:} led the research, designed and ran the scaling experiments, designed and experimented with the training strategies.
\textbf{JS,~TL,~EC,~AS,~WF,~XD:} advised the research, proposed experiments and helped with the writing.
\textbf{AS,~IB,~BZ:} ran preliminary experiments using label smoothing, longer training and RandAugment.
\textbf{IB:} demonstrated ResNets outperforming EfficientNets across all scales, designed the scaling strategies and the Pareto curve of models, designed/ran (semi-)supervised learning experiments and significantly contributed to the writing.
\textbf{BZ:} ran the regularization studies.
\textbf{WF,~BZ:} did a majority of the writing.
\textbf{BZ,~EC:} analyzed scaling experiments and generated the scaling plots.
\textbf{XD:} proposed, designed and ran the 3D video classification experiments, lead the open-sourcing.
\textbf{AS:} proposed lowering the weight decay for better performance and ran preliminary experiments comparing SimCLR to supervised learning.
\textbf{TL:} designed and ran the transfer learning experiments comparing to self-supervised learning.

\section{Details of all ResNet-RS models in the Pareto curve\label{sec:pareto_curve_details}}
This section details all the models in the ResNet-RS Pareto curve. 
In Table~\ref{tab:pareto_curve_details}, we observe that our ResNet-RS models get speedups ranging from \textbf{1.7x - 2.7x} across the EfficientNet Pareto curve on TPUs. 

\begin{table*}[h!]
\begin{center}
\small
\begin{tabular}{ccccccc}
  \toprule
  Model & Image Resolution & Params \textcolor{gray}{(M)} & FLOPs \textcolor{gray}{(B)} & V100 Latency \textcolor{gray}{(s)} & TPUv3 Latency \textcolor{gray}{(ms)} & Top-1 \\
  \midrule
  EfficientNet-B0 & 224 & 5.3 & 0.8 & 0.47 & 90 & 77.1 \\
  EfficientNet-B1 & 240 & 7.8 & 1.4 & 0.82 & 150 & 79.1 \\
  ResNet-RS-50 & 160 & 36 & 4.6 & 0.31 & 70 & 78.8\\
  \midrule
  
  EfficientNet-B2 & 260 & 9.2 & 2.0 & 1.03 & 210 & 80.1 \\
  ResNet-RS-101 & 160 & 64 & 8.4 & 0.48 \rlap{\textbf{\small{\textcolor{blue}{(2.1$\times$)}}}} & 120 \rlap{\textbf{\small{\textcolor{blue}{(1.8$\times$)}}}} & 80.3 \\
  \midrule
  
  EfficientNet-B3 & 300 & 12 & 3.6 & 1.76 & 340 & 81.6 \\
  ResNet-RS-101 & 192 & 64 & 12 & 0.70 & 170 & 81.2 \\ 
  ResNet-RS-152 & 192 & 87 & 18 & 0.99 & 240 & 82.0 \\
  \midrule
  
  EfficientNet-B4 & 380 & 19 & 8.4 & 4.0 & 710 & 82.9 \\
  ResNet-RS-152 & 224 & 87 & 24 & 1.48 \rlap{\textbf{\small{\textcolor{blue}{(2.7$\times$)}}}}& 320 \rlap{\textbf{\small{\textcolor{blue}{(2.2$\times$)}}}} & 82.8 \\
  ResNet-RS-152 & 256 & 87 & 31 & 1.76 \rlap{\textbf{\small{\textcolor{blue}{(2.3$\times$)}}}} & 410 \rlap{\textbf{\small{\textcolor{blue}{(1.7$\times$)}}}} & 83.0 \\
  \midrule
  
  EfficientNet-B5 & 456 & 30 & 20 & 8.16 & 1510 & 83.7 \\
  ResNet-RS-200 & 256 & 93 & 40 & 2.86 & 570 & 83.4 \\
  ResNet-RS-270 & 256 & 130 & 54 & 3.76 \rlap{\textbf{\small{\textcolor{blue}{(2.2$\times$)}}}} & 780 \rlap{\textbf{\small{\textcolor{blue}{(1.9$\times$)}}}} & 83.8 \\
  \midrule
  
  EfficientNet-B6 & 528 & 43 & 38 & 15.7 & 3010 & 84.0 \\
  ResNet-RS-350 & 256 & 164 & 69 & 4.72 \rlap{\textbf{\small{\textcolor{blue}{(3.3$\times$)}}}} & 1100 \rlap{\textbf{\small{\textcolor{blue}{(2.7$\times$)}}}} & 84.0 \\
  \midrule
  
  EfficientNet-B7 & 600 & 66 & 74 & 29.9 & 6020 & 84.7 \\
  ResNet-RS-350 & 320 & 164 & 107 & 8.48 & 1630 & 84.2 \\
  ResNet-RS-420 & 320 & 192 & 128 & 10.16 & 2090 & 84.4 \\
  \bottomrule
\end{tabular}
\end{center}
\caption{\textbf{Details of ResNet-RS models in Pareto curve.} 
All models are trained for 350 epochs using the improvements mentioned in Section~\ref{sec:training_setups}. 
The exact hyperparameters for all ResNet-RS models are in Table~\ref{tab:pareto_curve_hparams}.
Latencies on Tesla V100 GPUs are measured with full precision (\texttt{float32}).
Latencies on TPUv3 are measured using \texttt{bfloat16} precision.
All latencies are measured with an initial training batch size of 128 images, which is divided by 2 until it fits onto the accelerator.} 
\label{tab:pareto_curve_details} 
\end{table*}

\paragraph{Hyperparameters}
Table~\ref{tab:pareto_curve_hparams} presents the training and regularization hyperparameters used for training ResNet-RS models.
We increase regularization as with model scale.
Note that we have less hyperparameter setups compared to EfficientNets~\citep{tan2019efficientnet}.
We perform early stopping on the \texttt{minival-set} set for the two largest models from Table~\ref{tab:pareto_curve_details} (ResNet-RS-350 at resolution 320 and ResNet-RS-420 at resolution 320).
For every other model, we simply report the final accuracy.
We present top-1 accuracies on the ImageNet \texttt{test-set} for two ResNet-RS models in Table~\ref{tab:imagenet_test_split}.
We observe no sign of overfitting.

\begin{table}[h!]
    \small
    \begin{center}
    \begin{tabular}{ccccccc}
      \toprule
      Model & Depth & Image Resolution & RandAugment & Stochastic Depth & Dropout   \\
      & & & Magnitude & Rate & Rate \\
      \midrule
      ResNet-RS & 50 & 160 $\times$ 160 & 10 & 0.0 & 0.25\\
      \midrule
      
      ResNet-RS & 101 & 160 $\times$ 160 & 10 & 0.0 & 0.25\\
      \midrule
      
      ResNet-RS & 101 & 192 $\times$ 192 & 15 & 0.0 & 0.25\\ 
      ResNet-RS & 152 & 192 $\times$ 192 & 15 & 0.0 & 0.25\\ 
      ResNet-RS & 152 & 224 $\times$ 224 & 15 & 0.0 & 0.25\\ 
      \midrule
      
      ResNet-RS & 152 & 256 $\times$ 256 & 15 & 0.0 & 0.25\\ 
      \midrule
      
      ResNet-RS & 200 & 256 $\times$ 256 & 15 & 0.1 & 0.25\\ 
      ResNet-RS & 270 & 256 $\times$ 256 & 15 & 0.1 & 0.25\\ 
      \midrule
      
      ResNet-RS & 350 & 256 $\times$ 256 & 15 & 0.1 & 0.25\\
      \midrule
      
      ResNet-RS & 350 & 320 $\times$ 320 & 15 & 0.1 & 0.4\\
      ResNet-RS & 420 & 320 $\times$ 320 & 15 & 0.1 & 0.4\\
      \bottomrule
    \end{tabular}
    \end{center}
    \caption{\textbf{Hyperparameters for all ResNet-RS models.} 
    All models train for 350 epochs, use a weight decay of 4e-5, an EMA value of 0.9999 (for both weights and Batch Norm moving averages), 2 layers of RandAugment (with different magnitudes as shown above) and a label smoothing rate of 0.1. 
    The learning rate is warmed up to a maximum value of $0.1 / B$, with B the batch size, and decayed to 0 using a cosine schedule~\cite{loshchilov2016sgdr}.
    Dropout rate means each activation after the global average pooling layers gets dropped out with probability \emph{dropout rate}.
    }
\label{tab:pareto_curve_hparams} 
\end{table}

\begin{table}[h!]
\begin{center}
\small
\begin{tabular}{cccc}
  \toprule
  Model & Image Resolution & top-1 Val & top-1 Test \\
  \midrule
  ResNet-RS-152 & 224 & 82.8 & 82.7 \\
  ResNet-RS-270 & 256 & 83.8 & 83.7 \\
  \bottomrule
\end{tabular}
\end{center}
\vspace{-0.3cm}
\caption{\textbf{ImageNet accuracies on the validation and test splits.}} 
\label{tab:imagenet_test_split} 
\end{table}

\section{ResNet-RS Training and Regularization Methods\label{sec:appendix_training_details}}
Table~\ref{tab:hparam_comparison} shows the differences in training and regularization methods between ResNets, ResNet-RS and EfficientNets. Overall we closely match EfficientNet's training setup, while making a few minor simplications: cosine learning rate isntead of exponential decay and Momentum instead of RMSProp. 
Both simplifications reduce the total number of hyperparameters as \textbf{(1)} cosine decay has no hyperparameters associated with it and \textbf{(2)} Momentum has one less than RMSProp.

\begin{table*}[h!]
\begin{center}
\small
\begin{tabular}{cccc}
  \toprule
   & ResNet (2015) & ResNet-RS (2021) & EfficientNets (2019) \\
  \hline
  Epochs Trained & 90 & 350 & 350 \\
  LR Decay Schedule & Stepwise & Cosine & Exponential Decay \\
  
  Optimizer & Momentum & Momentum & RMSProp \\
  EMA of Weights & & \checkmark & \checkmark \\
  Label Smoothing & & \checkmark & \checkmark \\
  Stochastic Depth & & \checkmark & \checkmark \\
  RandAugment & & \checkmark & \checkmark \\
  Dropout on FC & & \checkmark & \checkmark \\
  Smaller Weight Decay & & \checkmark & \checkmark \\
  
  Squeeze-Excitation & & \checkmark & \checkmark \\
  Stem Modifications & & \checkmark & \checkmark \\
  \bottomrule
\end{tabular}
\end{center}
\caption{\textbf{Comparing training method between ResNet, ResNet-RS and EfficientNet.} ResNet (2015) refers to the ResNet originally trained in~\citet{resnet}.}
\label{tab:hparam_comparison} 
\end{table*}

\section{ResNet-RS Architecture Details\label{sec:architectural_details}}

\begin{table}[h!]
\begin{center}
\begin{tabular}{ccc}
  \toprule
  Model & Depth & Block Configuration   \\
  \hline
  ResNet & 50 & \texttt{[3-4-6-3]}\\
  ResNet & 101 & \texttt{[3-4-23-3]}\\
  ResNet & 152 & \texttt{[3-8-36-3]}\\ 

  ResNet & 200 & \texttt{[3-24-36-3]}\\ 
  ResNet & 270 & \texttt{[4-29-53-4]}\\
  ResNet & 350 & \texttt{[4-36-72-4]}\\
  ResNet & 420 & \texttt{[4-44-87-4]}\\
  \bottomrule
\end{tabular}
\end{center}
\caption{\textbf{Block configurations for all ResNet depths used in the ResNet-RS Pareto Curve.} ResNets of depths 50, 101, 152 and 200 use the standard block allocations from \citet{resnet}. The different numbers represent the number of blocks in \texttt{c2}, \texttt{c3}, \texttt{c4} and \texttt{c5} respectively. Note that our depth scaling mainly scales the blocks in \texttt{c3} and \texttt{c4}, which limits overfitting (due to the increase in parameters) that can occur when blocks are added to \texttt{c5}.}
\label{tab:block_config_details} 
\end{table}

\begin{figure}[h!]
    \begin{center}
    \includegraphics[width=0.3\linewidth]{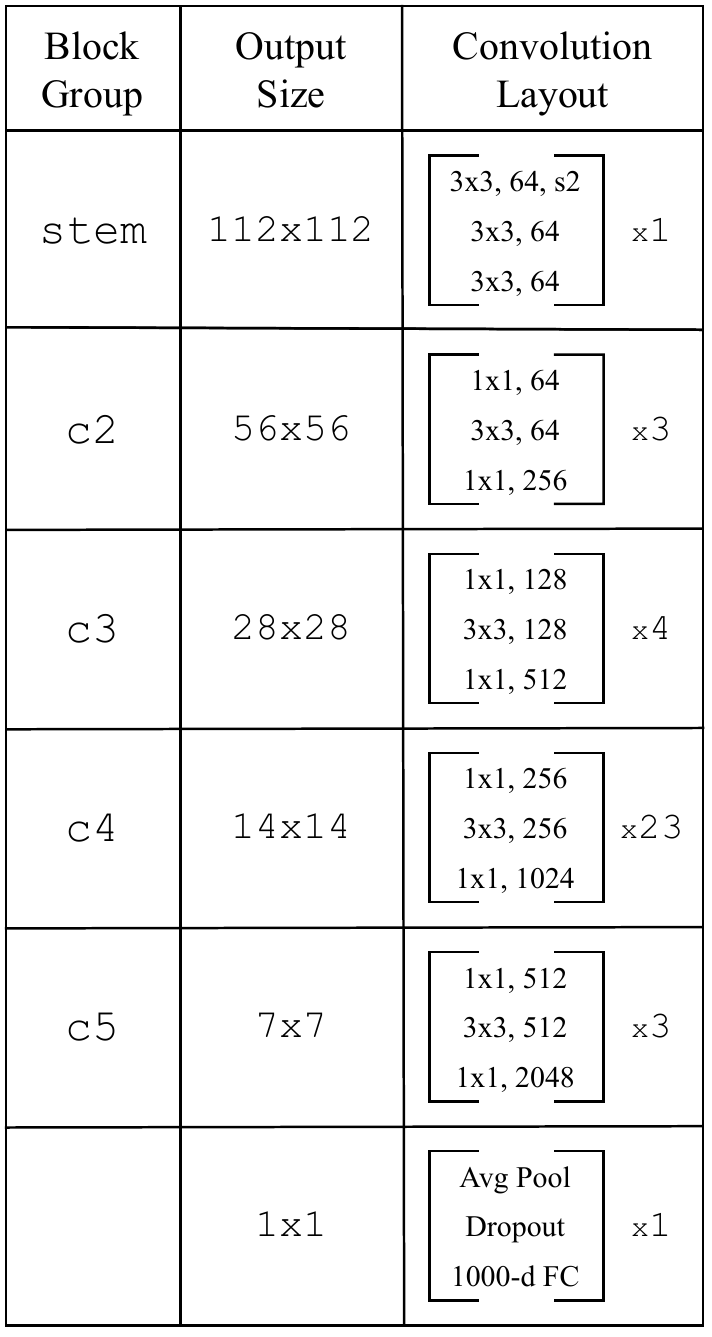}
    \end{center}
    \caption{\textbf{ResNet-RS Architecture Diagram.} Output Size assumes a 224$\times$224 input image resolution. In the convolutional layout column $x2$ refers to the the first $3\times3$ convolution being applied with a stride of 2. The ResNet-RS architecture is a simple combination of Squeeze-and-Excitation and ResNet-D. The $\times$ symbol refers to how many times the blocks are repeated in the ResNet-101 architecture. These values change across depths according to the blocks layouts in Table~\ref{tab:block_config_details}.}
    \label{fig:resnet_arch_diagram}
\end{figure}

We provide more details of the ResNet-RS architectural changes. 
We reiterate that ResNet-RS is a combination of: improved scaling strategies, improved training methodologies, the ResNet-D modifications~\cite{he2019bag} and the Squeeze-Excitation module~\cite{hu2018squeeze}.
Table~\ref{tab:block_config_details} shows the block layouts for all ResNet depths used throughout our work. ResNet-50 through ResNet-200 use the standard block configurations from~\citet{resnet}. 
ResNet-270 and onward primarily scale the number of blocks in \texttt{c3} and \texttt{c4} and we try to keep their ratio roughly constant. 
We empirically found that adding blocks in the lower stages limits overfitting as blocks in the lower layers have significantly less parameters, even though all blocks have the same amount of FLOPs. Figure~\ref{fig:resnet_arch_diagram} shows the ResNet-D architectural changes used in our ResNet-RS models.

\section{Scaling Analysis Regularization and Model Details}
\label{sec:reg_scheme_scaling}
\begin{table}[ht!]
\centering
\begin{tabular}{c|c}
  \toprule
  Filter Scaling  & Dropout Rate \\
  \hline
  0.25 & 0.0  \\
  0.5 & 0.1 \\
  1.0 & 0.25 \\
  1.5 & 0.6 \\
  2.0 & 0.75\\
  \bottomrule
\end{tabular}
\caption{\textbf{Dropout values for filter scaling.} Filter scaling refers to the filter scaling multiplier based on the number of filters in the original ResNet architecture. }
\label{tab:dropout_scaling_values} 
\end{table}

\paragraph{Regularization for 350 epoch models.} 
The dropout rates used for various filter multipliers (across all image resolutions and depths) are in Table~\ref{tab:dropout_scaling_values}. 
RandAugment is used with 2 layers and its magnitude is set to 10 for filter multipliers in [0.25, 0.5] or image resolution in [64, 160], 15 for image resolution in [224, 320] and 20 otherwise.
We apply stochastic depth with a drop rate of 0.2 for image resolutions 224 and above.
We do not apply stochastic depth filter multiplier 0.25 (or images smaller than 224).
All models use a label smoothing of 0.1 and a weight decay of 4e-5. 
These values were set based on the preliminary experiments across various model scales on the ImageNet \texttt{minival-set}.

\paragraph{Regularization for 10 and 100 epochs.} 
We did not use RandAugment, Dropout, Stochastic Depth or Label Smoothing. Flips and crops were used and a weight decay of 4e-5.

\textbf{Block allocation for ResNet-300 and ResNet-400.} 
For ResNet 101 and ResNet-200 we use the block allocations decribed in Table~\ref{tab:block_config_details}. For ResNet-300, our block allocation is \texttt{[4-36-54-4]} and ResNet-400 is \texttt{[6-48-72-6]}.

\section{Fine-Tuning Protocols}
For fine-tuning we initialize the parameters in the ResNet backbone with a pre-trained model and randomly initialize the rest of the layers. 
We perform \textit{end-to-end} fine-tuning with an extensive grid search of the combinations of learning rate and training steps to ensure each pre-trained model achieves its best fine-tuning performance. 
We experiment with different weight decays but do not find it making a big difference and set it to 1e-4. All models are trained with cosine learning rate for simplicity. 
Below we describe the dataset, evaluation metric, model architecture, and training parameters for each task.
\paragraph{CIFAR-100:} We use standard CIFAR-100 train and test sets and report the top-1 accuracy. We resize the image resolution to $256\times256$. We replace the classification head in the pre-trained model with a randomly initialized linear layer that predicts 101 classes, including background. We use a batch size of 512 and search the combination of training steps from 5000 to 20000 and learning rates from 0.005 to 0.32. We find the best learning rate for SimCLR (0.16) is much higher than SimCLRv2 (0.01) and the supervised model (0.005). This trend holds for the following tasks.

\paragraph{PASCAL Segmentation:} We use PASCAL VOC 2012 train and validation sets and report the mIoU metric. The training images are resampled into $512\times512$ with scale jittering [0.5, 2.0] (i.e. randomly resample image between $256\times256$ to $1024\times1024$ and crop it to $512\times512$). We remove the classification head and add randomly initialized FPN~\cite{fpn} layers. We follow the practice in~\cite{zoph2020rethink} to combine $P_3$ to $P_7$ and upsample it to $P_2$. The segmentation head consists of 3 convolution layers after $P_2$ layer and a linear layer to predict 21 categories including background at each pixel location. We use a batch size of 64 and search the combination of training steps from 5000 to 20000 and learning rates from 0.005 to 0.32.

\paragraph{PASCAL Detection:}
We use PASCAL VOC 2007+2012 trainval set and VOC 2007 test set and report the $AP_{50}$ with 11 recall points to compute average precision. The training images are resampled into $896\time896$ with scale jittering [0.5, 2.0]. We remove the classification head and add randomly initialized FPN~\cite{fpn} layers from $P_3$ to $P_7$. We use Faster R-CNN~\cite{ren2015faster} consisting a region proposal head and a \texttt{4conv1fc} Fast R-CNN head. We use a batch size of 32 and search the combination of training steps from 5000 to  20000 and learning rates from 0.005 to 0.32.

\paragraph{NYU Depth:} We use NYU depth v2 dataset with 47584 train and 654 validation images. We report the percentage of predicted depth values within 1.25 relative ratio compared to the ground truth. The training images are resampled into $640\time640$ with scale jittering [0.5, 2.0]. The model architecture is identical to segmentation model, except the last linear layer predicts a single depth value per pixel. We use a batch size of 64 and search the combination of training steps from 10000 to 40000 and learning rates from 0.005 to 0.32.

\section{Video Classification Experimental Details\label{sec:video_classification_details}}
We follow the training and inference protocols in~\cite{Qian2020SpatiotemporalCV, Feichtenhofer2019SlowFastNF}.
We train with a random 224$\times$224 crop or its horizontal flip on the spatial domain and sample a 32-frame clip with temporal stride 2. 
We use a $1024$ batch size, $0.8$ learning rate with cosine decay and train for $200$ epochs for the baseline.
At inference, we use 256$\times$256 crop size for the spatial domain and adopt the 30 views protocol~\cite{Feichtenhofer2019SlowFastNF}. 

Starting from the baseline, we apply the following training methods: dropout with a rate of $0.5$, $0.1$ label smoothing, stochastic depth with $0.2$ drop rate, EMA of weights, smaller weight decay (set to 4e-5) and a $350$ epoch training schedule.
For data augmentation, we use scale jittering~\cite{Qian2020SpatiotemporalCV} as a replacement to RandAugment. 
We adjust the stochastic depth rate to $0.1$ when applying scale jittering to optimize performance.
To implement the ResNet-D stem for the 3D ResNet, we use the same kernel configurations for the spatial domain and use temporal kernel sizes of $[5,1,1]$ for the three layers.

\section{Profiling Setup\label{sec:profiling}}
All latencies refer to training latencies.
All models were run on TPUv3~\cite{jouppi2017datacenter} with \texttt{bfloat16} precision in TensorFlow 1.x.
TPU latencies are measured on 8 TPUv3 cores with a batch size of 1024 (i.e. 128 per core) which is divided by 2 until it fits onto the accelerator's memory.
In the cases where a smaller batch size is employed, we normalize the reported latency to the original batch size of 1024 images.
For GPU profiling we use a single Tesla-V100 with \texttt{float32} precision with a starting batch size of 128, also divided by multiples of 2 if necessary.

\end{document}